\documentclass[10pt,twocolumn,letterpaper]{article}
\usepackage{cvpr} 

\usepackage[accsupp]{axessibility}
\usepackage{times}
\usepackage{epsfig}
\usepackage{graphicx}
\usepackage{verbatim}
\usepackage{amsmath}
\usepackage{amssymb}
\usepackage{colortbl, color, xcolor}
\usepackage{makecell}

\usepackage[pagebackref=true,breaklinks=true,colorlinks,bookmarks=false]{hyperref}

\usepackage{amstext} 
\usepackage{array}   
\newcolumntype{L}{>{$}c<{$}} 

\newcommand{\fref}[1]{Figure~\ref{#1}}
\newcommand{\tref}[1]{Table~\ref{#1}}

\def\cvprPaperID{****} 

\usepackage{cleveref}

\begin{document}

\title{NTIRE 2025 Image Shadow Removal Challenge Report}

\author{
Florin-Alexandru Vasluianu$^\dagger$ \and Tim Seizinger$^\dagger$ \and Zhuyun Zhou$^\dagger$ \and Cailian Chen$^\dagger$ \and Zongwei Wu$^\dagger$ \and Radu Timofte$^\dagger$ \and
Mingjia Li \and
Jin Hu \and
Hainuo Wang \and
Hengxing Liu \and
Jiarui Wang \and
Qiming Hu \and
Xiaojie Guo \and
Xin Lu \and
Jiarong Yang \and
Yuanfei Bao \and
Anya Hu \and
Zihao Fan \and
Kunyu Wang \and
Jie Xiao \and
Xi Wang \and
Xueyang Fu \and
Zheng-Jun Zha \and
Yu-Fan Lin \and
Chia-Ming Lee \and
Chih-Chung Hsu \and
Xingbo Wang \and
Dong Li \and
Yuxu Chen \and
Bin Chen \and
Yuanbo Zhou \and
Yuanbin Chen \and
Hongwei Wang \and
Jiannan Lin \and
Qinquan Gao \and
Tong Tong \and
Zhao Zhang \and
Yanyan Wei \and
Wei Dong \and
Han Zhou \and
Seyed Amirreza Mousavi  \and
Jun Chen \and
Haobo Liang \and
Jiajie Jing \and
Junyu Li \and
Yan Yang \and
Seoyeon Lee \and
Chaewon Kim \and
Ziyu Feng \and
Shidi Chen \and
Bowen Luan \and
Zewen Chen \and
Vijayalaxmi Ashok Aralikatti \and
G Gyaneshwar Rao \and
Nikhil Akalwadi \and
Chaitra Desai \and
Ramesh Ashok Tabib \and
Uma Mudenagudi \and
Anas M. Ali \and
Bilel Benjdira \and
Wadii Boulila \and
Alexandru Brateanu \and
Cosmin Ancuti \and
Tanmay Chaturvedi \and
Manish Kumar \and
Anmol Srivastav \and
Daksh Trivedi \and
Shashwat Thakur \and
Kishor Upla \and
Zeyu Xiao \and
Zhuoyuan Li \and
Boda Zhou \and
Shashank Shekhar \and
Kele Xu \and
Qisheng Xu \and
Zijian Gao \and
Tianjiao Wan \and
Suiyi Zhao \and
Bo Wang \and 
Yan Luo \and
Mingshen Wang \and
Yilin Zhang
}

\maketitle


\begin{abstract}
This work examines the findings of the NTIRE 2025 Shadow Removal Challenge.
A total of 306 participants have registered, with 17 teams successfully submitting their solutions during the final evaluation phase.
Following the last two editions, this challenge had two evaluation tracks: one focusing on reconstruction fidelity and the other on visual perception through a user study. 
Both tracks were evaluated with images from the WSRD+ dataset, simulating interactions between self- and cast-shadows with a large number of diverse objects, textures, and materials. 
\end{abstract}
{\let\thefootnote\relax\footnotetext{%
\hspace{-5mm} 
$\dagger$
Florin-Alexandru Vasluianu, Tim Seizinger, 
Zhuyun Zhou, Cailian Chen, Zongwei Wu, 
and Radu Timofte are the NTIRE 
2025
challenge organizers. The other authors participated in the challenge.
\\
Appendix \ref{sec:apd:team} (in the supplementary) provides names and affiliations.
\\ 
\url{https://cvlai.net/ntire/2025}
}}

\section{Introduction}
Shadow removal can be considered as a particular sub-task towards the broader field of Image Restoration, having gained significant research interest due to the complexity of the shadow formation process.
The intensity and shape of shadows are influenced by a large number of factors, such as the light source, the geometry of the occluding object, and the properties of the recipient surface.
The intricacy of this system demands comprehensive investigation, with Shadow Removal and detection persisting as an area of active research.

Early works rely on the physical properties of a determined shadow formation model, which is usually characterized by a restricted set of conditions.
One category of Shadow Removal solutions using classical image processing aims to restore shadow-free images by transferring local statistics from non-shadowed to shadowed areas \cite{finlayson2002removing, Finlayson_entropyminimization}.
While promising, such methods require high-quality shadow detection as a pre-processing task to define shadow areas, which constitutes an equally difficult problem.
However, solutions built on localization information regarding the shadow affected areas remain a popular design choice~\cite{10.1145/1243980.1243982,Shor:2008:TSM,7893803}.

Powered by increasingly faster hardware, later solutions employed deep learning paradigms to exploit large-scale image databases for shadow removal and/or detection~\cite{Gryka2015softShadows, SRDDESHADOW, ISTDwang2018STCGAN, USRhu2019mask, vasluianu2023shadow}.
These solutions are adopting fully-supervised \cite{ISTDwang2018STCGAN, Le_2019_ICCV, Le_2020_ECCV, SRDDESHADOW, zhang2022spa, vasluianu2023shadow, guo2023shadowformer}, weakly-supervised or self-supervised~\cite{USRhu2019mask, ding2019argan, vasluianu2021shadow, jin2021dc} training frameworks. 

Recently, diffusion models \cite{ho2020denoising} have been increasingly adopted by various image shadow removal solutions \cite{guo2023shadowdiffusion, jin2024des3}.
These approaches are characterized by the highest perceptual quality in the currently established benchmarks, such as ISTD\cite{ISTDwang2018STCGAN}, ISTD+\cite{ISTDwang2018STCGAN, Le_2020_ECCV}, or SRD\cite{SRDDESHADOW}.

All aforementioned solutions advanced shadow removal, but the shadow formation model studied by them is limited to specific and simplified scenarios.
LRSS\cite{Gryka2015softShadows} offers soft shadow images, while SRD\cite{SRDDESHADOW} and ISTD\cite{ISTDwang2018STCGAN} are limited to scenarios with hard shadows.
Their acquisition setups are defined with natural sun light as the only light source, with an occluding object that itself is not visible in the scene, occluding parts thereof in the input image.
The reference image is then obtained by simply moving the light occluder away from the scene so that no shadow is cast onto it.
Due to the duration of this procedure and variability of natural light due to, e.g. moving clouds, these pairs can show significant variance in color balance.
Although a color correction method was introduced in \cite{Le_2020_ECCV}, defining the improved variations of ISTD+ and SRD+, various semantic inconsistencies still exist in these datasets \cite{vasluianu2021shadow, jin2021dc}.

The main limitation of the setup used to acquire the aforementioned datasets~\cite{Gryka2015softShadows, SRDDESHADOW, ISTDwang2018STCGAN} is that the areas affected by the shadows are always flat with little geometric complexity. 
Naturally, this is because more complex scene geometries would result in self-shadowing being included in the reference frame.
However, self-shadows represent the most prevalent form of natural shadows; thus, their study is imperative for practical applications beyond scenario-specific prototypes.

To alleviate these limitations, the authors of WSRD~\cite{vasluianu2023shadow} adopt a professional photography studio setup, using artificial light sources for both input and reference acquisition.
While a directional point light source is used to simulate natural sunlight that affects the scene, a omnidirectional softbox-based light setup was used to acquire reference images.
This setup enables near-optimal lighting distribution, even when rough surfaces or complex scene geometry are present.
Therefore, WSRD can represent complex self-shadowing phenomena, narrowing the gap between the dataset and the in-the-wild domains.

Recently, the WSRD setup was extended \cite{vasluianu2024towards}, with multiple directional lights of varying sizes, positions, and distance.
This more complex setup thereby is modeling varying intensities in the umbra-shadow region and varying transitions within the penumbra between shadow and non-shadow affected areas.
This advancement broadens the scope of the shadow-removal task by introducing an image database specifically to address the connected Ambient Lighting Normalization problem.



This challenge is one of the NTIRE 2025~\footnote{\url{https://www.cvlai.net/ntire/2025/}} Workshop associated challenges on: ambient lighting normalization~\cite{ntire2025ambient}, reflection removal in the wild~\cite{ntire2025reflection}, shadow removal, event-based image deblurring~\cite{ntire2025event}, image denoising~\cite{ntire2025denoising}, XGC quality assessment~\cite{ntire2025xgc}, UGC video enhancement~\cite{ntire2025ugc}, night photography rendering~\cite{ntire2025night}, image super-resolution (x4)~\cite{ntire2025srx4}, real-world face restoration~\cite{ntire2025face}, efficient super-resolution~\cite{ntire2025esr}, HR depth estimation~\cite{ntire2025hrdepth}, efficient burst HDR and restoration~\cite{ntire2025ebhdr}, cross-domain few-shot object detection~\cite{ntire2025cross}, short-form UGC video quality assessment and enhancement~\cite{ntire2025shortugc,ntire2025shortugc_data}, text to image generation model quality assessment~\cite{ntire2025text}, day and night raindrop removal for dual-focused images~\cite{ntire2025day}, video quality assessment for video conferencing~\cite{ntire2025vqe}, low light image enhancement~\cite{ntire2025lowlight}, light field super-resolution~\cite{ntire2025lightfield}, restore any image model (RAIM) in the wild~\cite{ntire2025raim}, raw restoration and super-resolution~\cite{ntire2025raw}, and raw reconstruction from RGB on smartphones~\cite{ntire2025rawrgb}.

\section{Challenge Data}

The WSRD dataset \cite{vasluianu2023shadow}, used in the NTIRE 2023 Image Shadow Removal Challenge \cite{vasluianu2023ntire}, marked a significant advancement in studying complex shadow interactions by incorporating more intricate surfaces and diverse content.
However, its main drawback stemmed from the fact that the data capturing setup is partly mobile. 
Therefore, inconsistencies in the alignment of pixels have proved to consume significant efforts on the participants side, with complex strategies that alleviate these issues dominating the competition.

To address this, we implemented a double-stage image alignment strategy, in which the preliminary image alignment implements a keypoint matching system for homography estimation. 
Differences in illumination for severely underexposed image segments prove to contribute to a challenging environment for a small subset of outlier image pairs.
Therefore, we introduced a refined image alignment as a gradient descent affinity estimation based on object contour geometry as a second alignment step, further enhancing the quality of the data provided for the NTIRE 2025 Image Shadow Removal Challenge.  

This set of adjustments resulted in a notable increase in PSNR of approximately 2 dB for the baseline data, opening the path for enhanced quality results.
Thus, the received submissions presented higher-quality restored images, with improvements evident in both reconstruction fidelity and perceived image quality. This enhanced alignment allowed participating teams to concentrate more on refining their shadow removal algorithms. 

Furthermore, due to the constraints of the Codalab server used for validation \cite{codalab_competitions}, the size of both the validation and testing split was limited to a set of 75 novel image pairs, with contents partly disjoint from those represented in the training split.  The selection process ensured that the most challenging samples were retained without altering the overall distribution of represented scenarios, content, or, crucially, previously unseen training objects or surfaces \footnote{ \url{https://github.com/fvasluianu97/WSRD-DNSR}}.

\section{Evaluation}
This challenge was a double-track competition with distinct ranking criteria for each track. The following are the evaluation criteria.
\begin{enumerate}
    \item The reconstruction fidelity in terms of PSNR;
    \item The Structured Similarity Index (SSIM) \cite{wang2004image} score;
    \item The LPIPS \cite{zhang2018perceptual} distance between the restorations and the ground-truth images. We used the ImageNet pretrained AlexNet \cite{10.1145/3065386} for LPIPS feature extraction;
    \item The Mean Opinion Score (MOS) for the submitted predictions;
    \item An efficiency metric based on the number of parameters reported by each team. 
\end{enumerate}

\noindent \textbf{Fidelity Track: }The Fidelity Track ranks images using the well established quantitative metrics of PSNR, SSIM, and LPIPS with equal weight attributed to each. 
For solutions of similar fidelity, lower complexity approaches are favored through a parameter efficiency metric.

\noindent \textbf{Perceptual Track: }In the Perceptual Track, the principal metric used was the mean opinion score (MOS), derived from a user study conducted after all submissions have been received.

\section{Challenge Phases}
\begin{enumerate}
    \item \textbf{Development phase:} In this phase, participants were given access to the task description, along with a set of 1000 pairs of images to train their models.
    \item \textbf{Validation phase:} To validate their solutions, participants received a set of 75 input images input images from the WSRD+ validation split, with the ground truths remaining private. To enable measurement of validation performance, a Codalab server \cite{codalab_competitions} was set up, comparing submission images uploaded by each user with private reference images. 
    \item \textbf{Final phase:} A subset of 75 input images from the testing split of WSRD+ was sent to the challenge participants, with validation performed using the Codalab server. Test-set fine-tuning was restricted by limiting user submissions. Finally, a submission template was provided, with instructions for the final submission preparation.
    Each team providing a method description, the corresponding codes, information regarding the team members, affiliations, and the final set of 75 restored images, corresponding to the testing split inputs.
    
\end{enumerate}

\section{User Study}
For the perceptual track, the primary ranking criteria of the challenge was the Mean Opinion Score (MOS). 
The score was based on a user study where imaging experts, including professional photographers, analyzed the images. 
The utilized grading system ranges from 1 (worst) to 10 (best) and is defined such that the input image corresponds to a grade of 3. 
Grades 1 and 2 are kept for situations in which the algorithm degrades the input image without visible improvement in the shadow-affected areas. 
Meanwhile, successfully restored images obtained a rating ranging from 3 to 10, with increments of 0.5. 

The user study was carried out on a subset of challenging samples from the challenge test split, with the index $i$ in the subset $i \in \{3,4,15,17,27,39,41,55,70,73\}$ given the 0-based indexing in the test sample split. 
The analyzed images were chosen given the complexity of the shadow scenario, with contents both seen and unseen during training.

\section{Challenge Results}
The challenge ended with 17 valid submissions.  Section \ref{sec:methods} provides details about each of the solutions ranked in the final phase.

\begin{figure*}[h!]
    \centering
    \renewcommand{\arraystretch}{0.7}
    \setlength{\tabcolsep}{1pt}
    \newcommand{\scale}{0.162}
    \begin{tabular}{cccccc}
         Inputs & FuShaRem & ACVLab & LUMOS  & X-Shadow & Reference \\

         \includegraphics[width=\scale\linewidth]{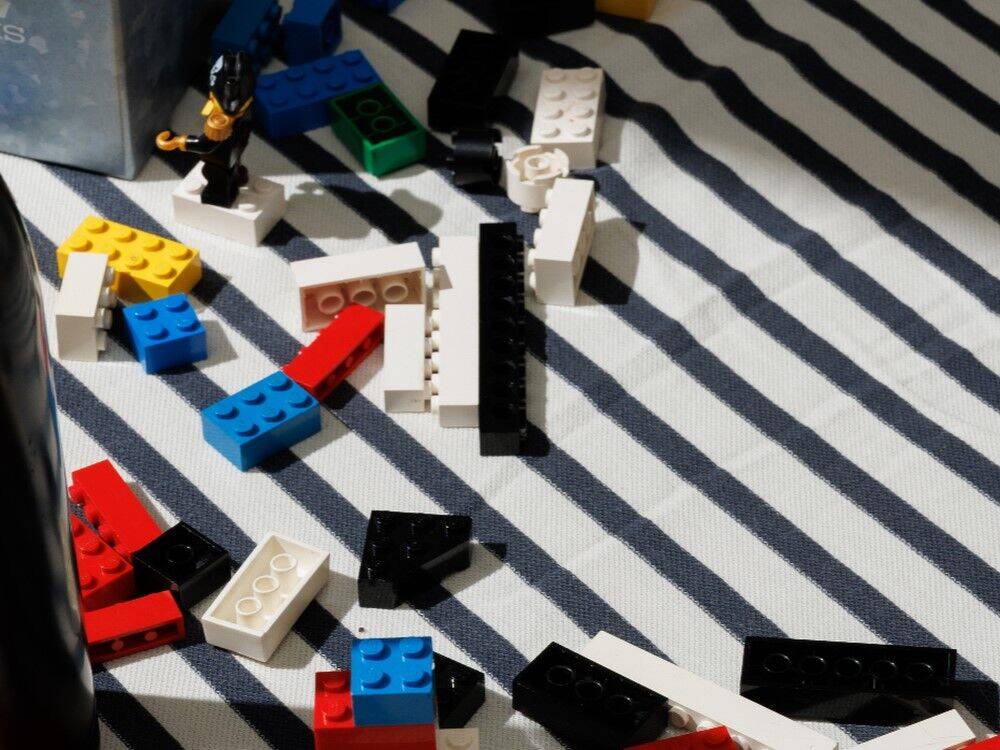} & 
         \includegraphics[width=\scale\linewidth]{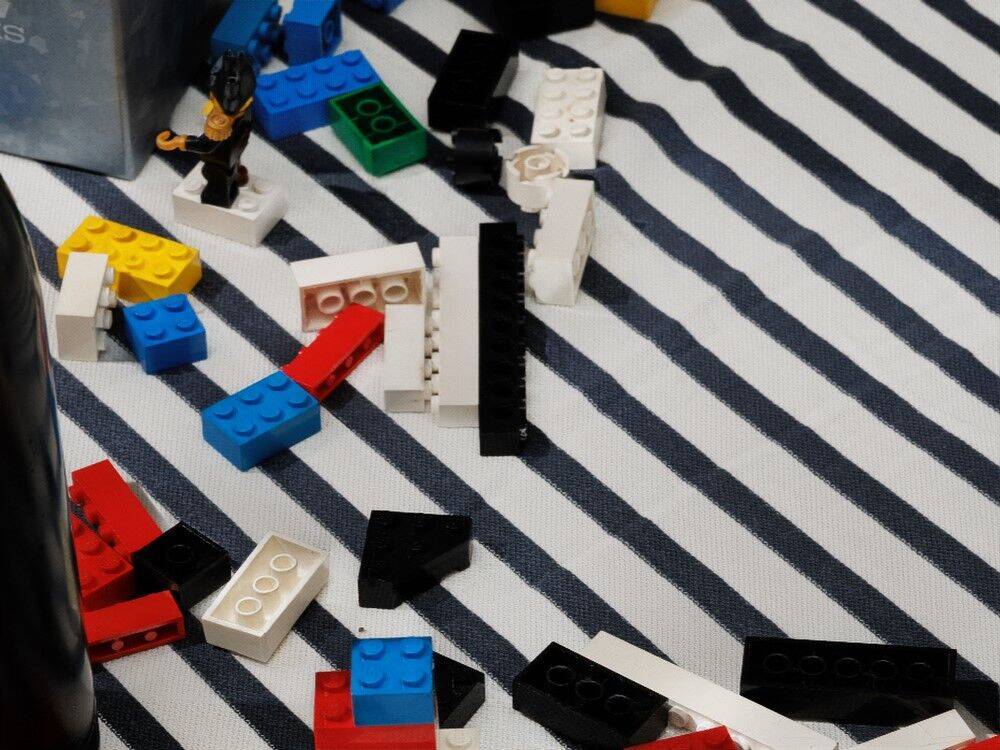} & 
         \includegraphics[width=\scale\linewidth]{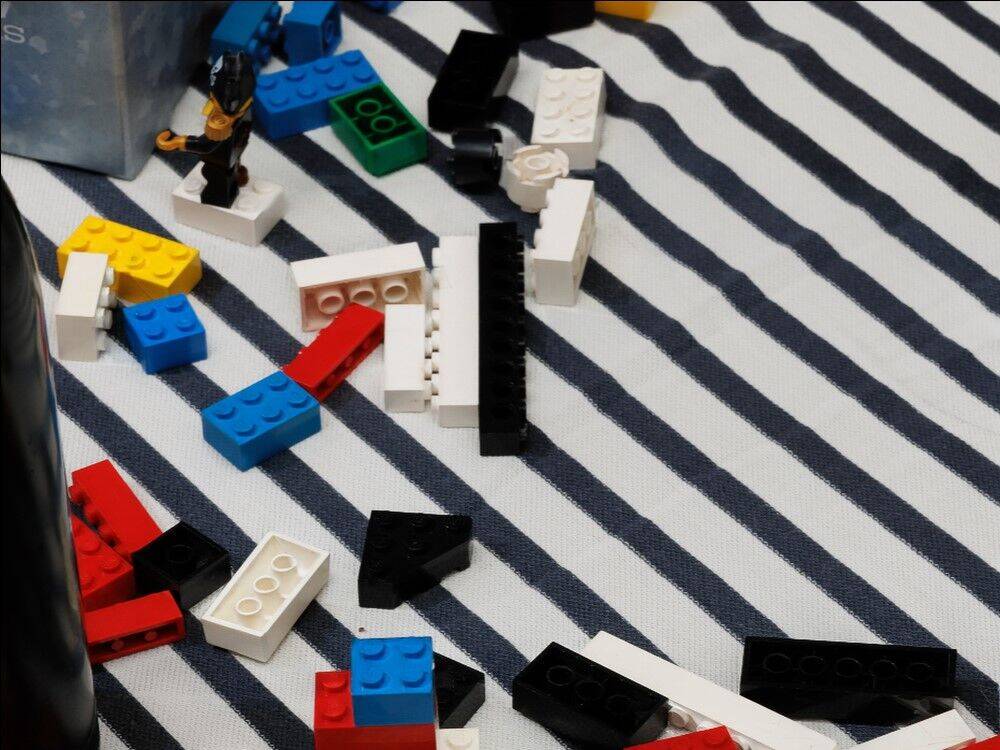} & 
         \includegraphics[width=\scale\linewidth]{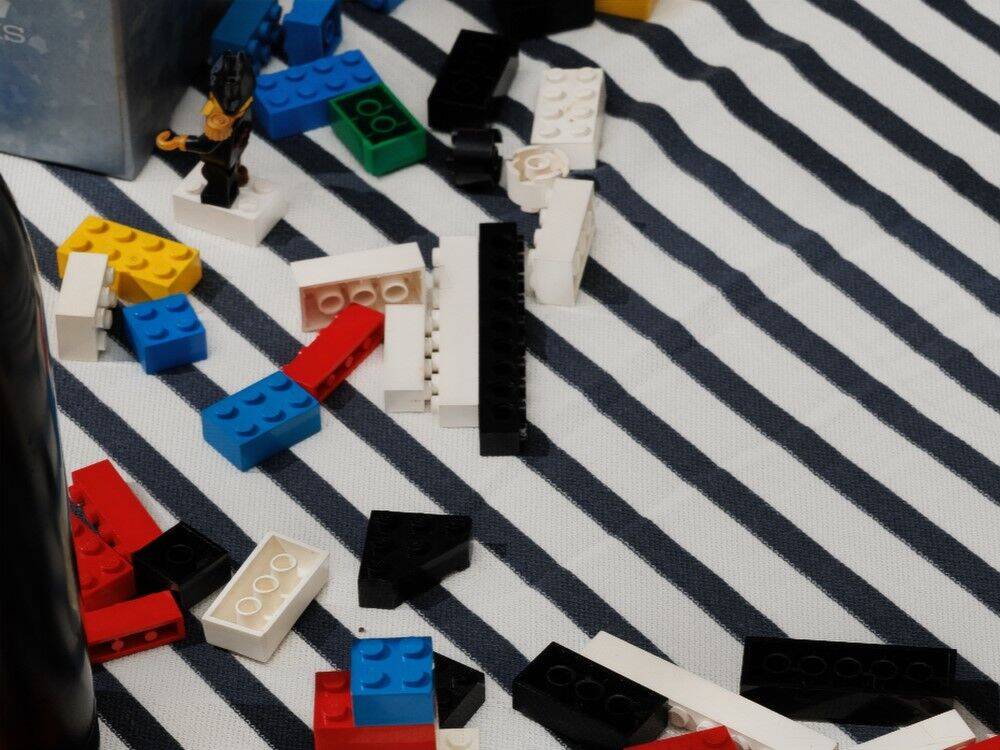} & 
         \includegraphics[width=\scale\linewidth]{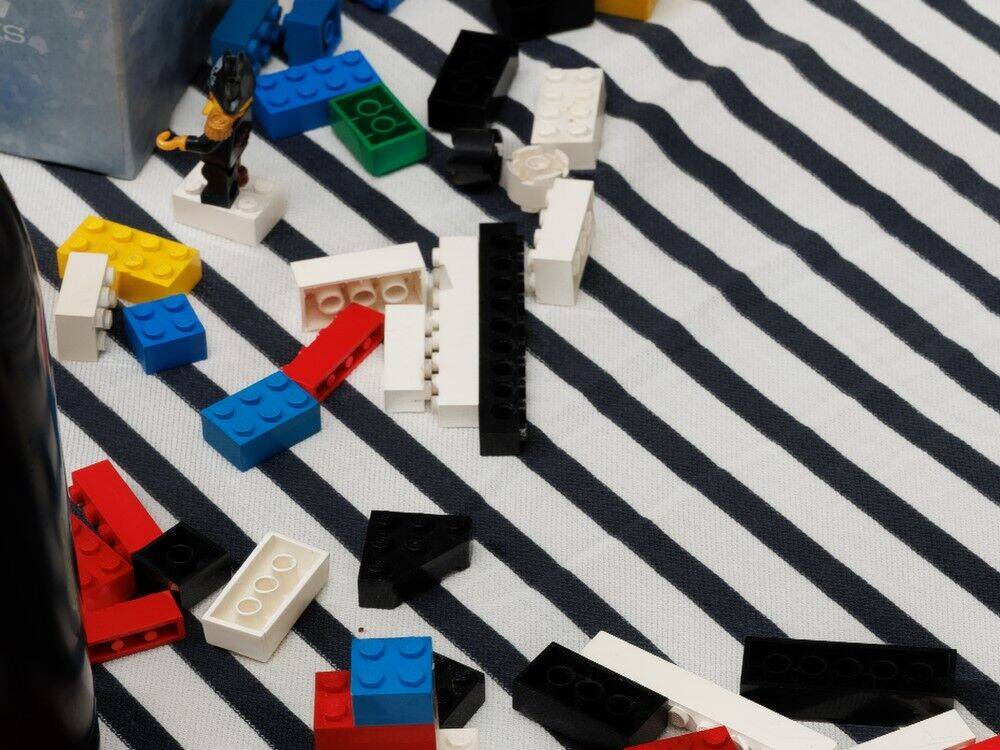} & 
         \includegraphics[width=\scale\linewidth]{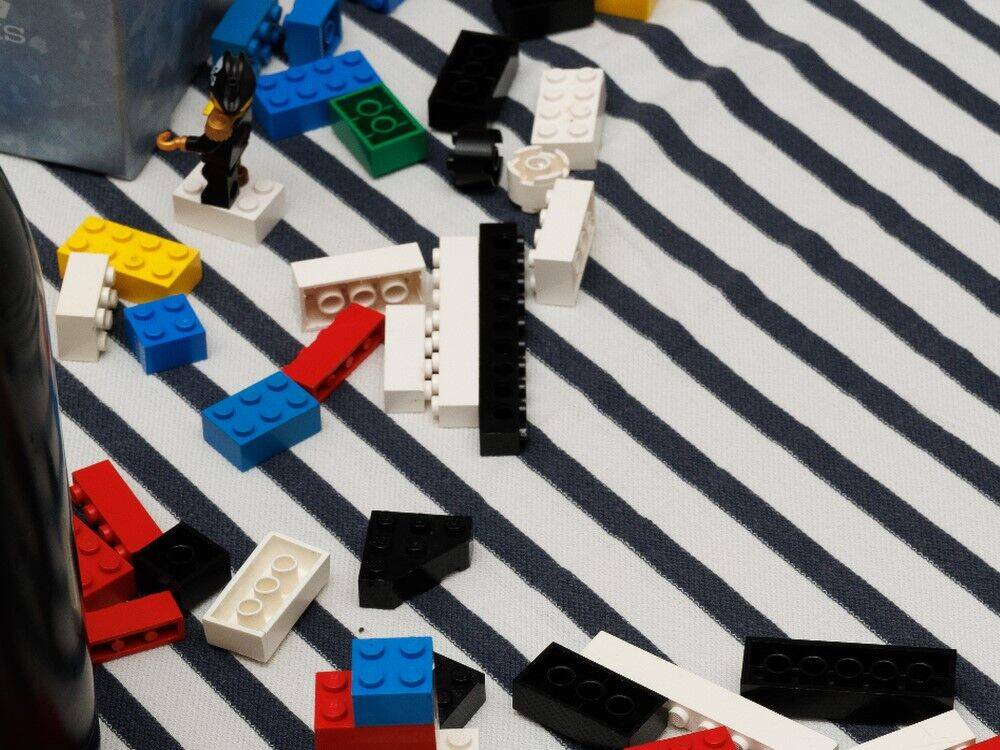} \\

         \includegraphics[width=\scale\linewidth]{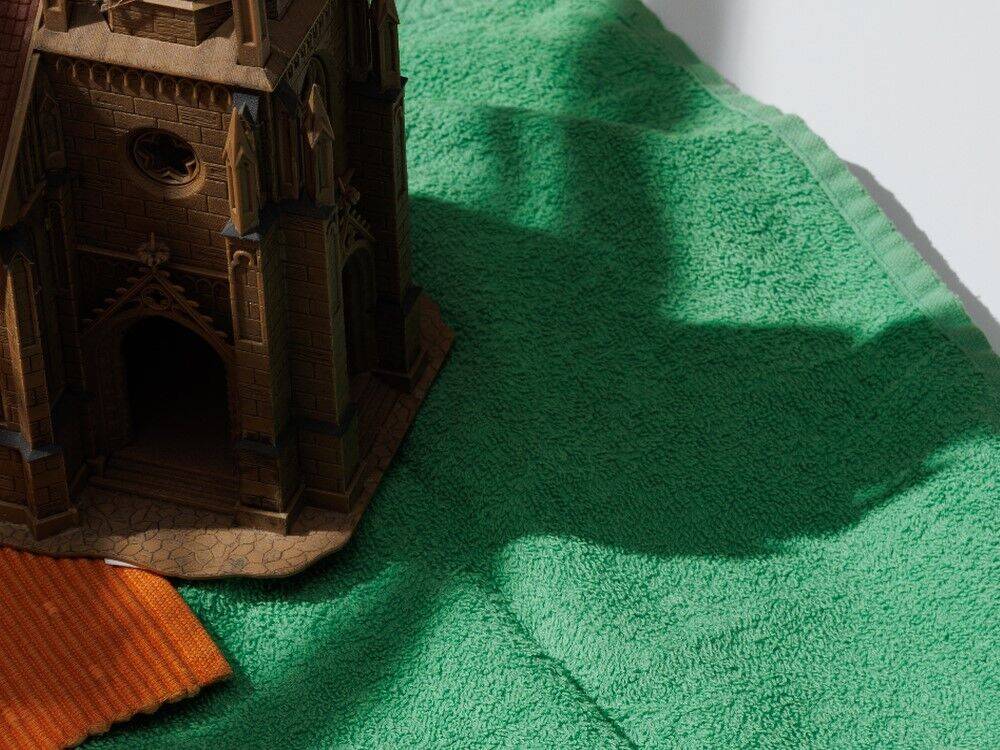} & 
         \includegraphics[width=\scale\linewidth]{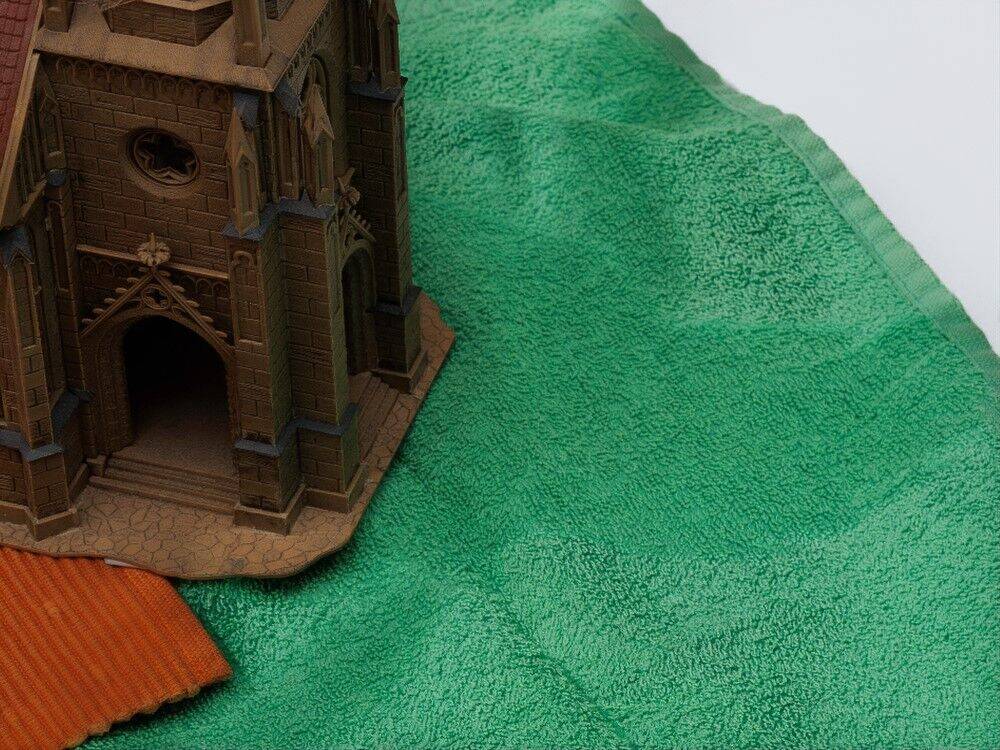} & 
         \includegraphics[width=\scale\linewidth]{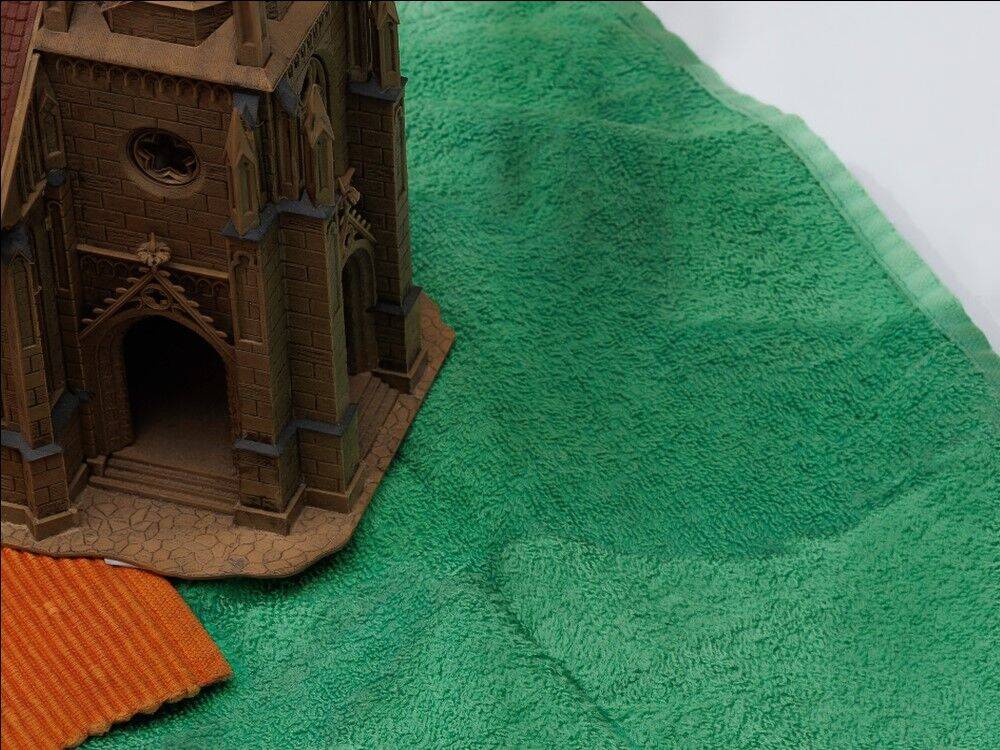} & 
         \includegraphics[width=\scale\linewidth]{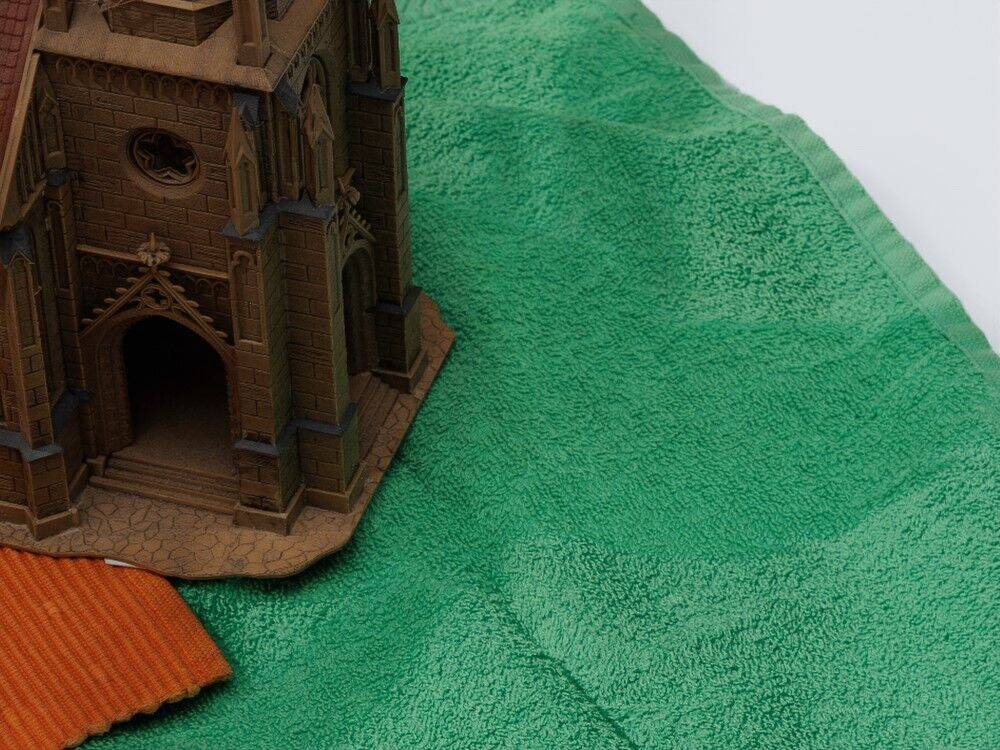} & 
         \includegraphics[width=\scale\linewidth]{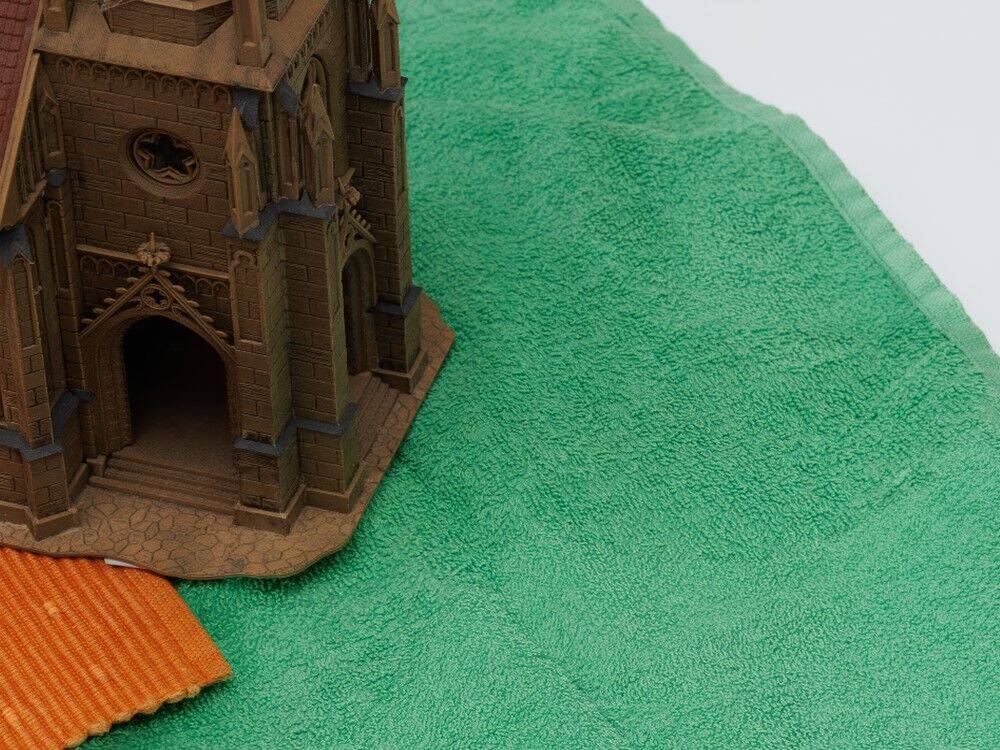} & 
         \includegraphics[width=\scale\linewidth]{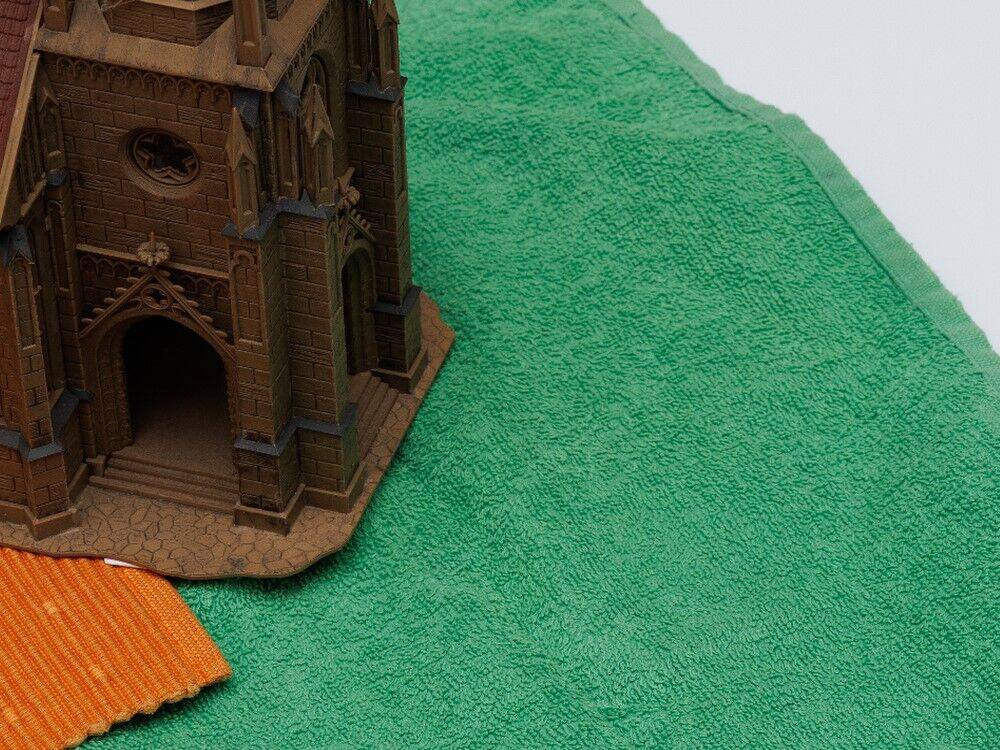} \\
         
         \includegraphics[width=\scale\linewidth]{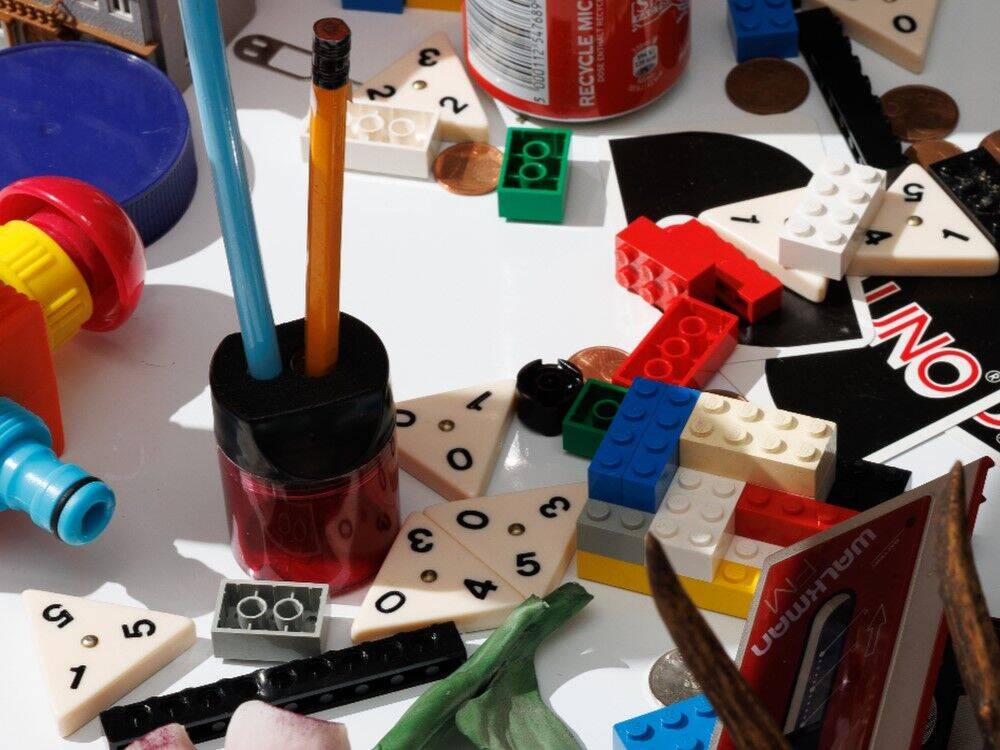} & 
         \includegraphics[width=\scale\linewidth]{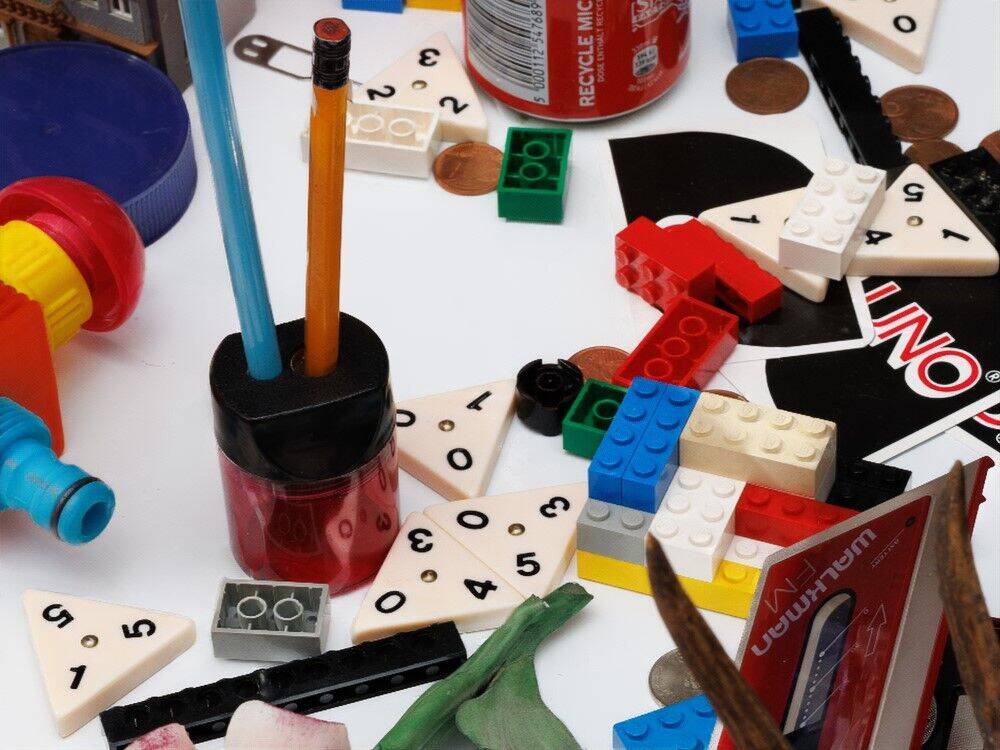} & 
         \includegraphics[width=\scale\linewidth]{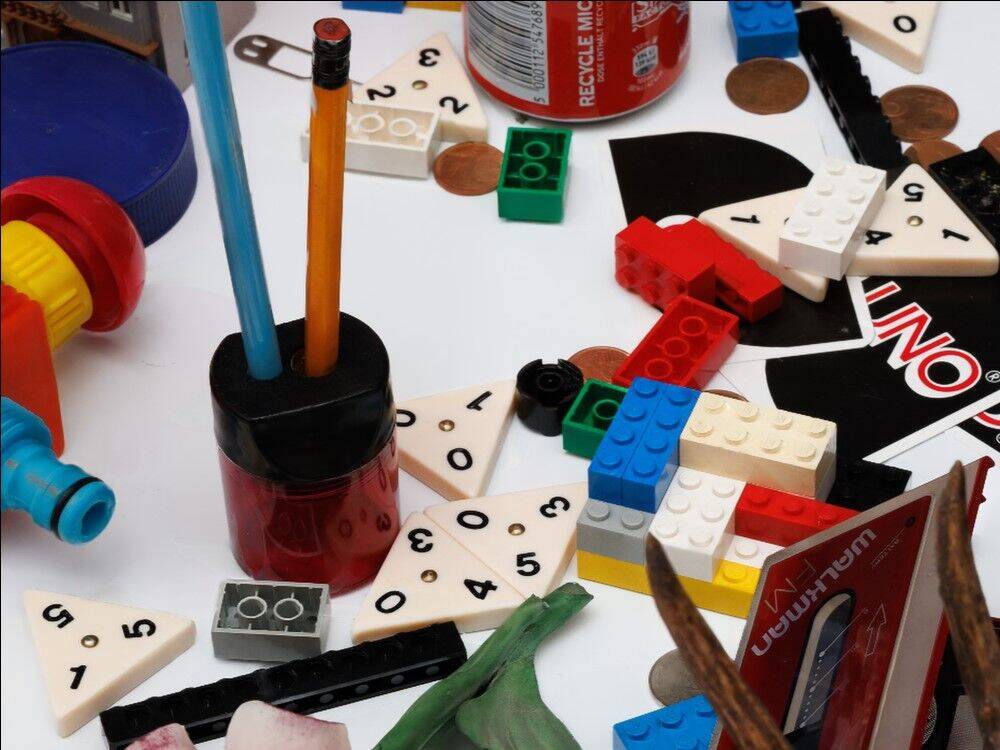} & 
         \includegraphics[width=\scale\linewidth]{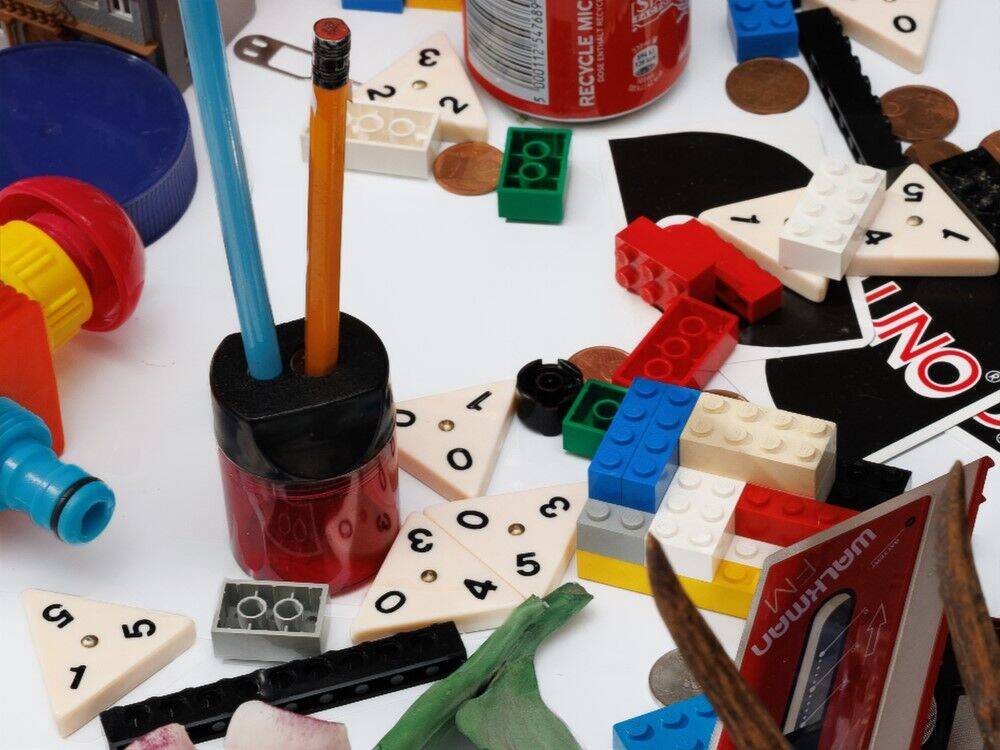} & 
         \includegraphics[width=\scale\linewidth]{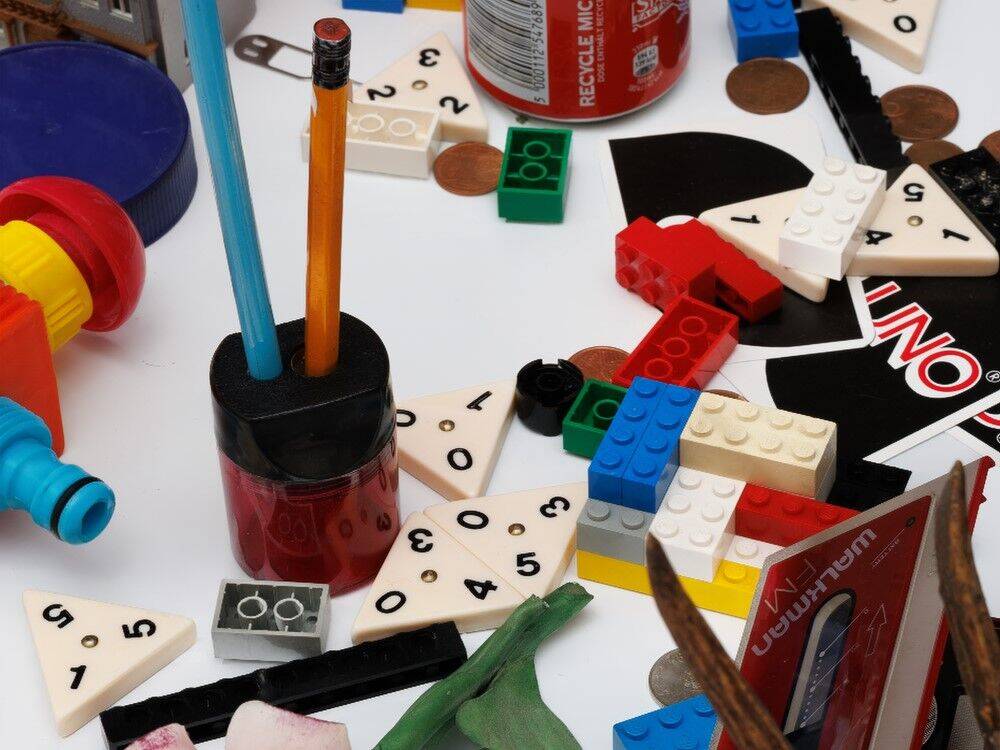} & 
         \includegraphics[width=\scale\linewidth]{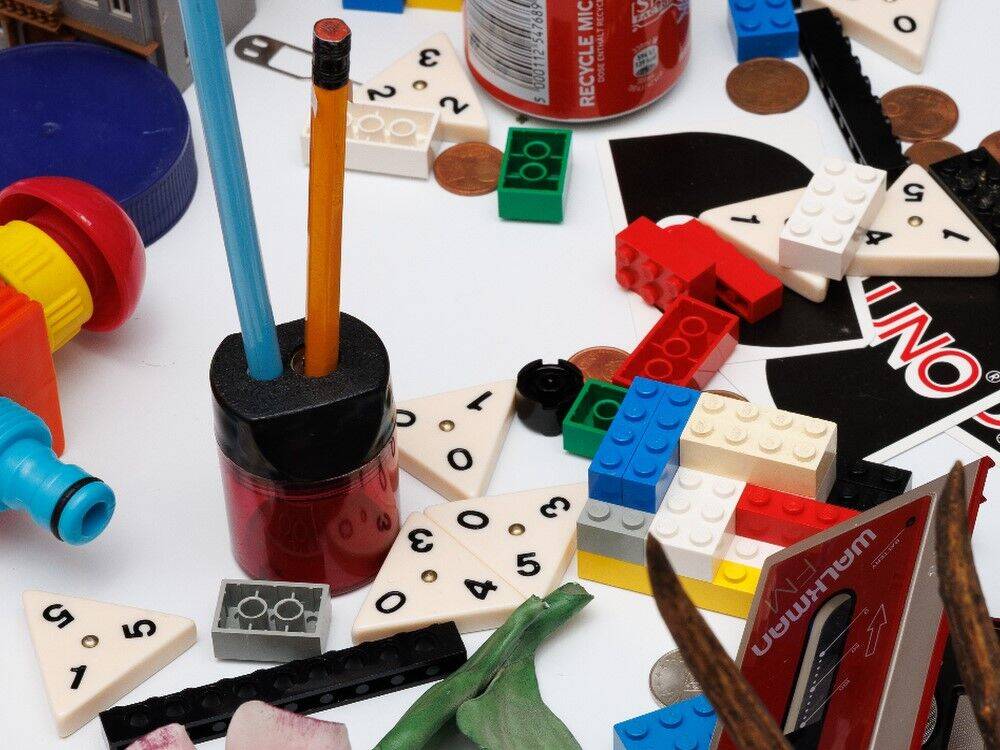} \\
         
         \includegraphics[width=\scale\linewidth]{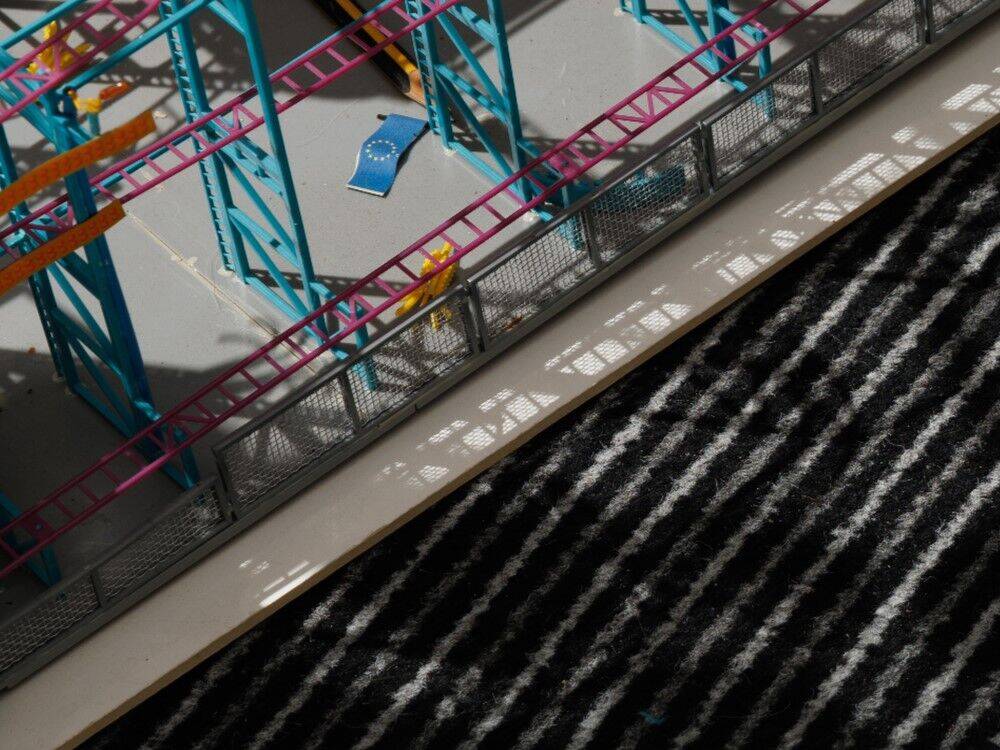} & 
         \includegraphics[width=\scale\linewidth]{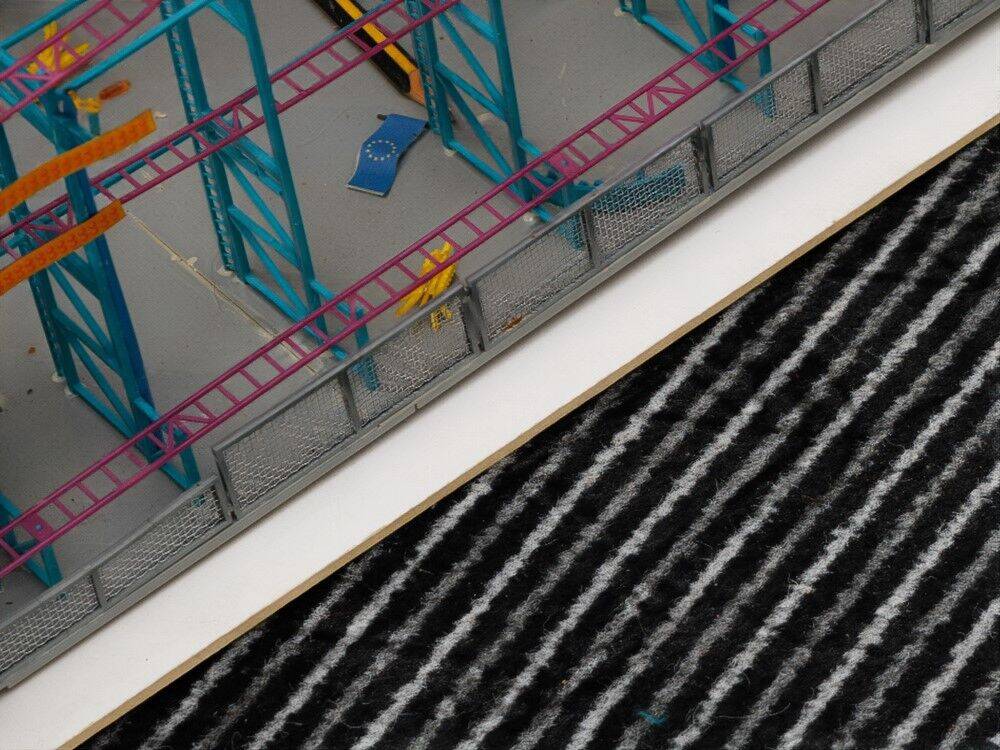} & 
         \includegraphics[width=\scale\linewidth]{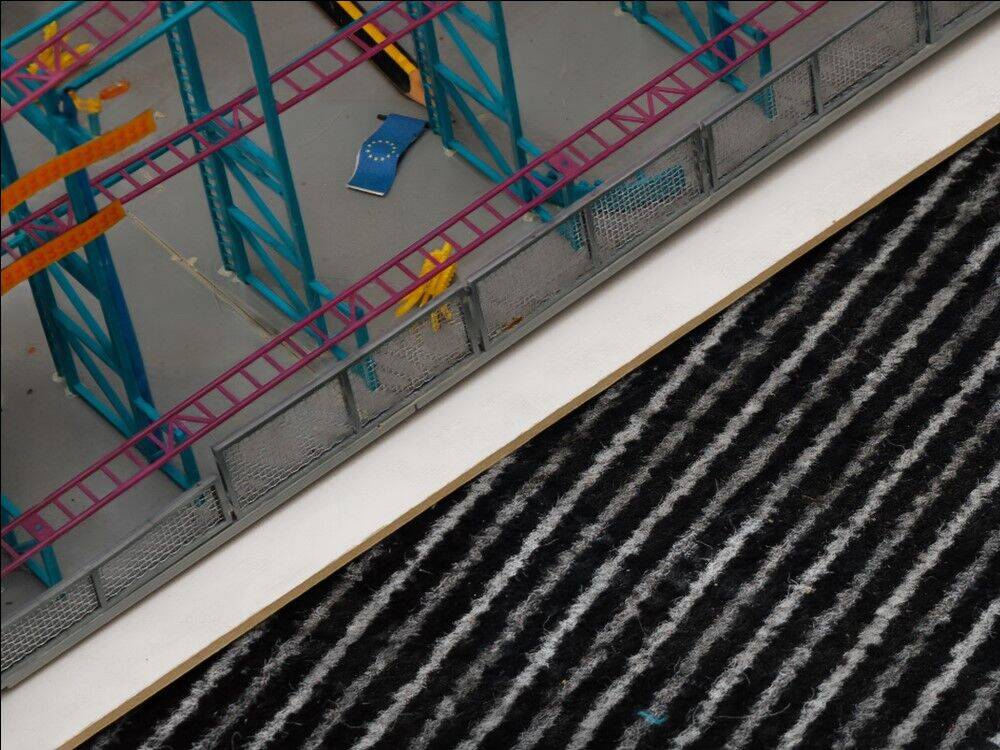} & 
         \includegraphics[width=\scale\linewidth]{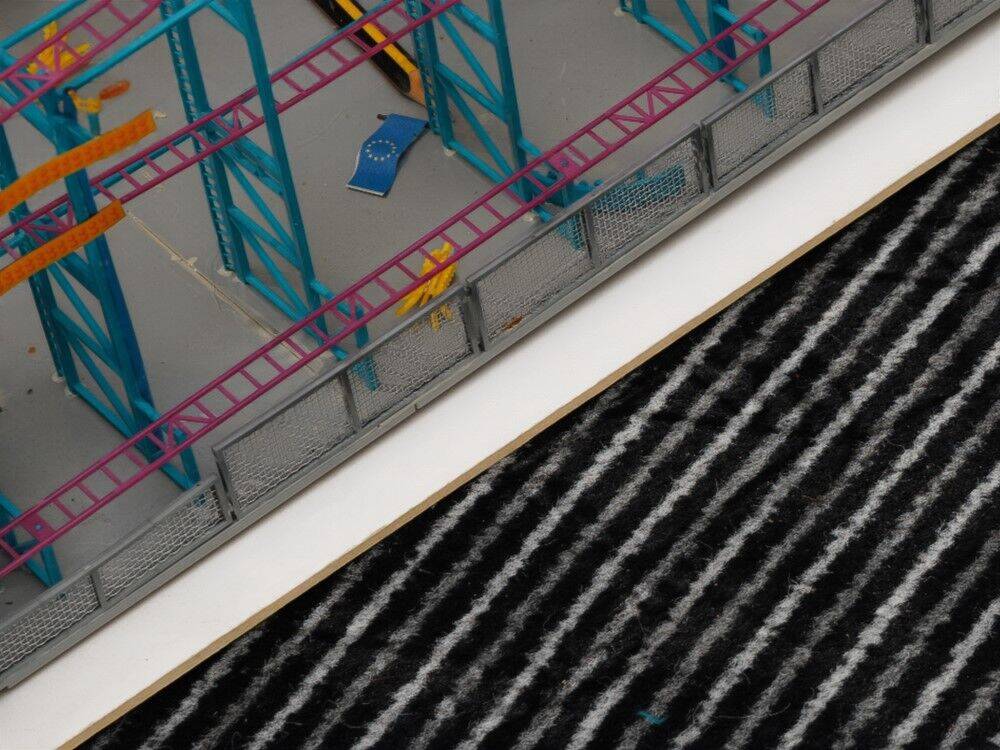} & 
         \includegraphics[width=\scale\linewidth]{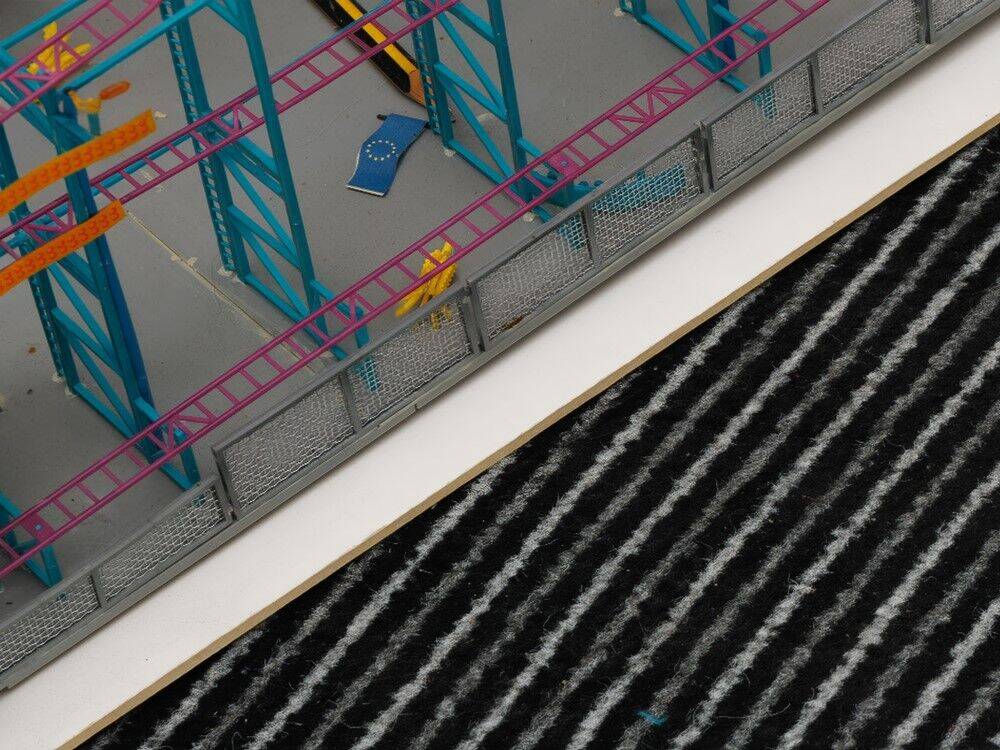} & 
         \includegraphics[width=\scale\linewidth]{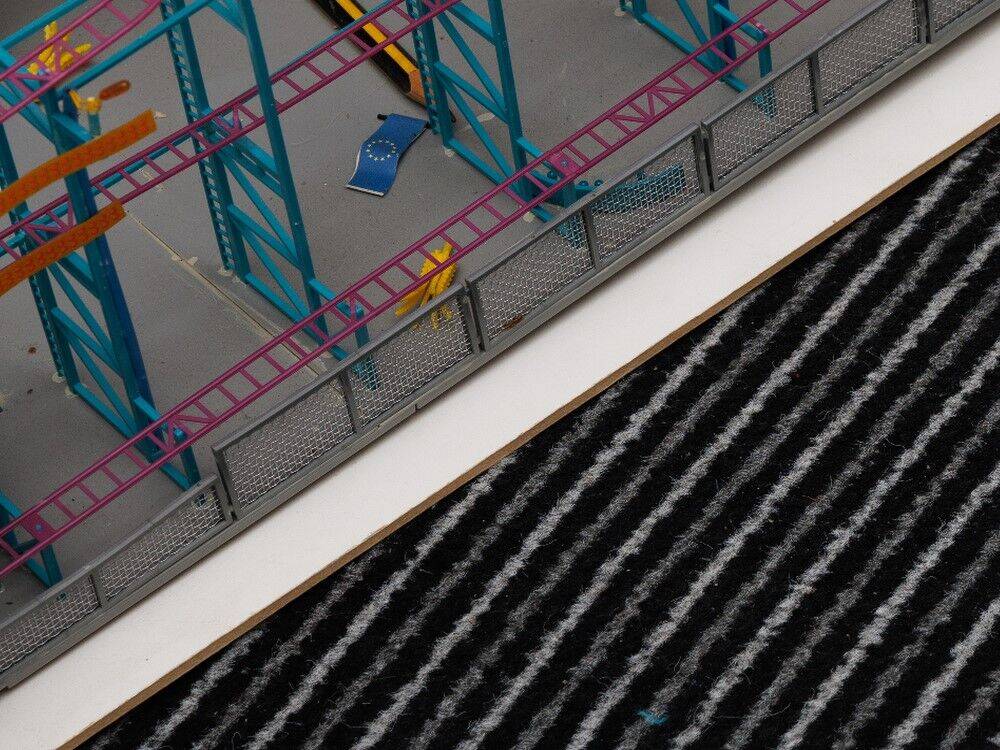} \\

         \includegraphics[width=\scale\linewidth]{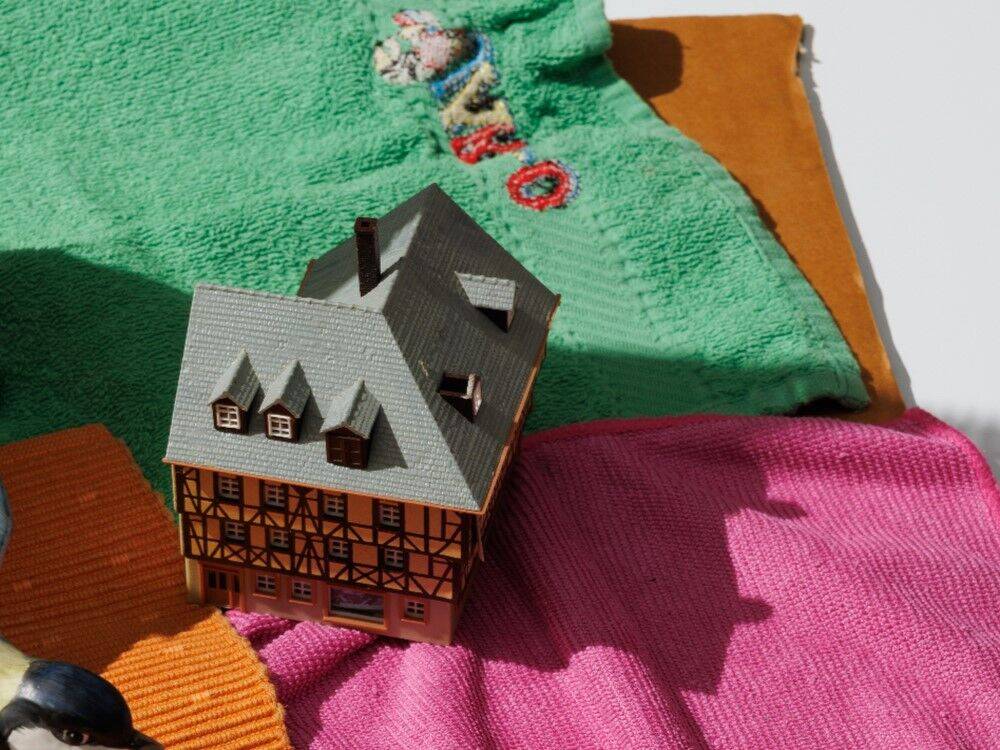} & 
         \includegraphics[width=\scale\linewidth]{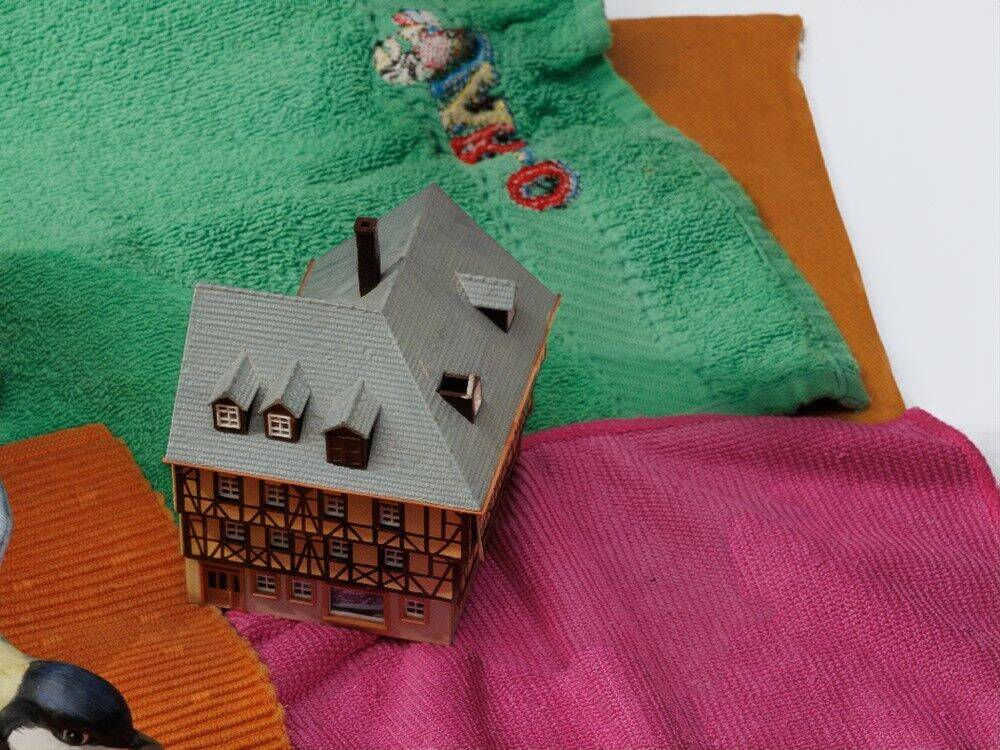} & 
         \includegraphics[width=\scale\linewidth]{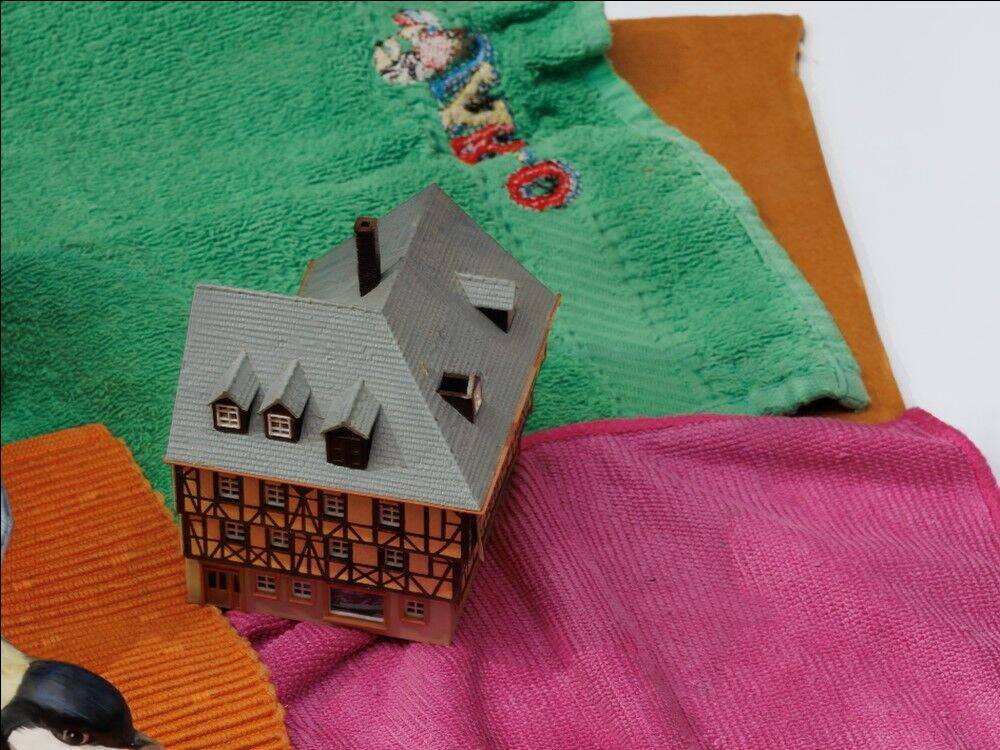} & 
         \includegraphics[width=\scale\linewidth]{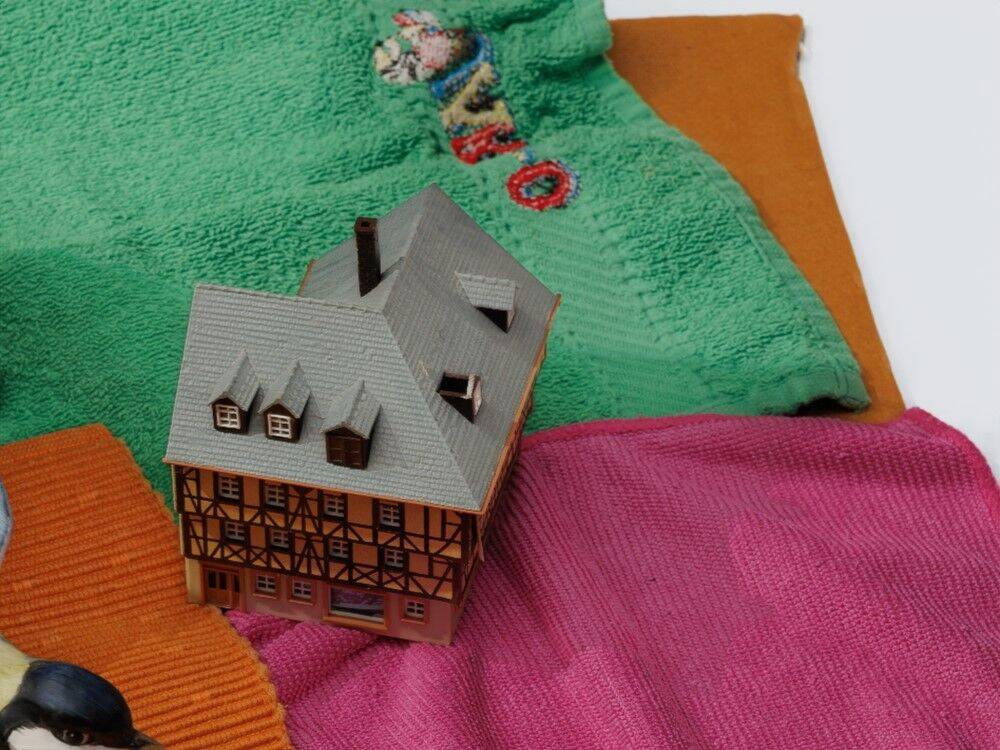} & 
         \includegraphics[width=\scale\linewidth]{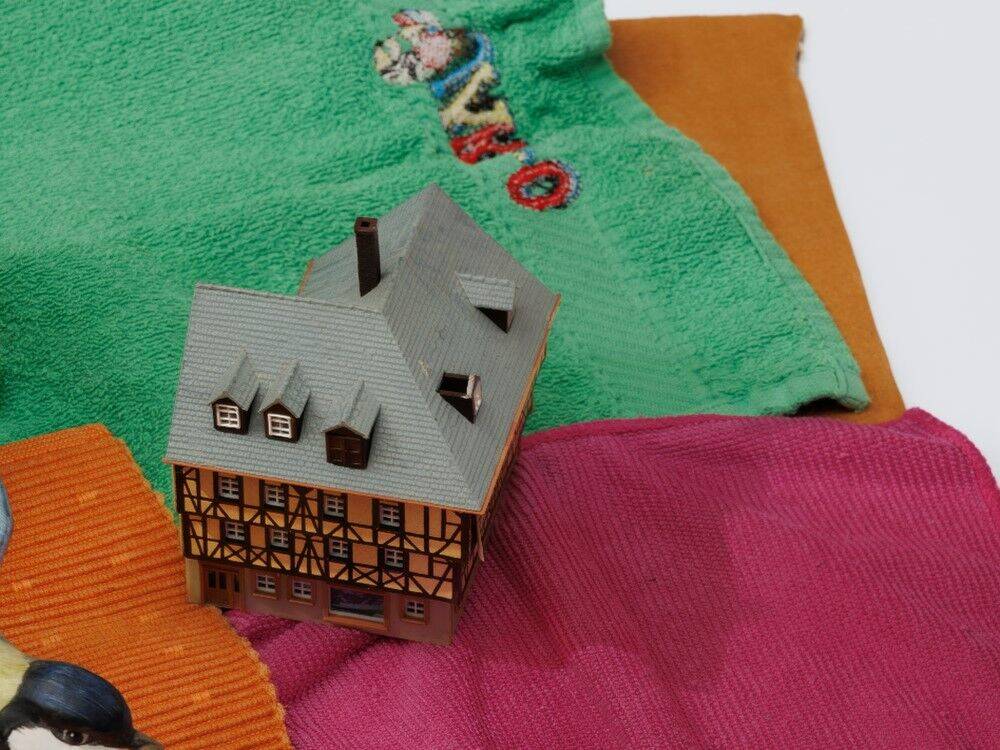} & 
         \includegraphics[width=\scale\linewidth]{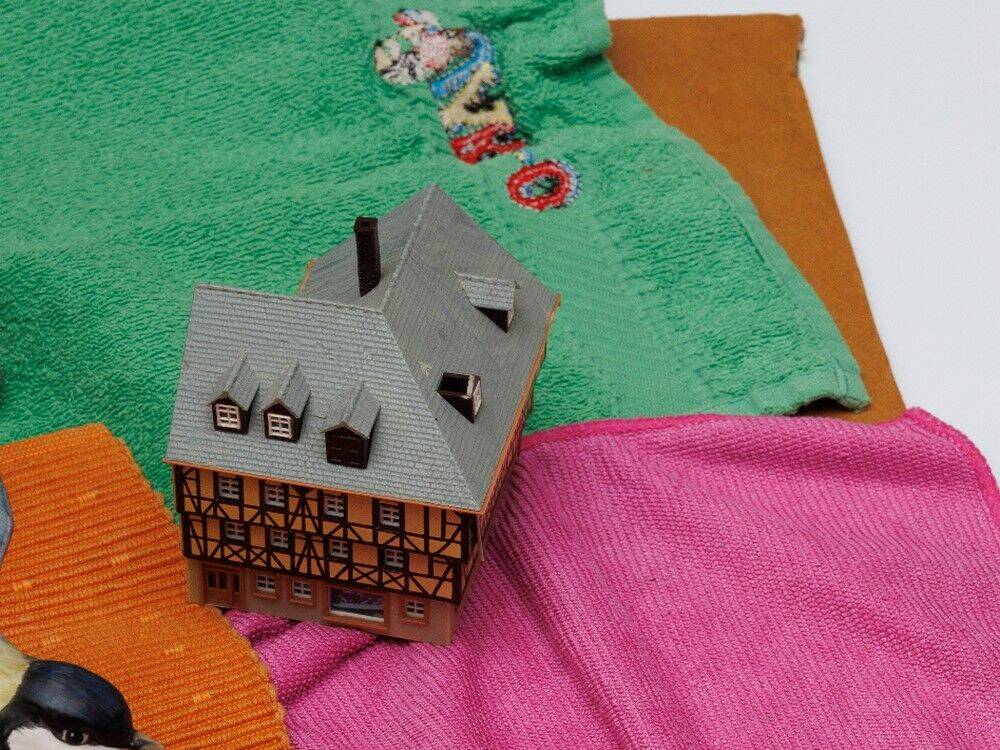} \\

         \includegraphics[width=\scale\linewidth]{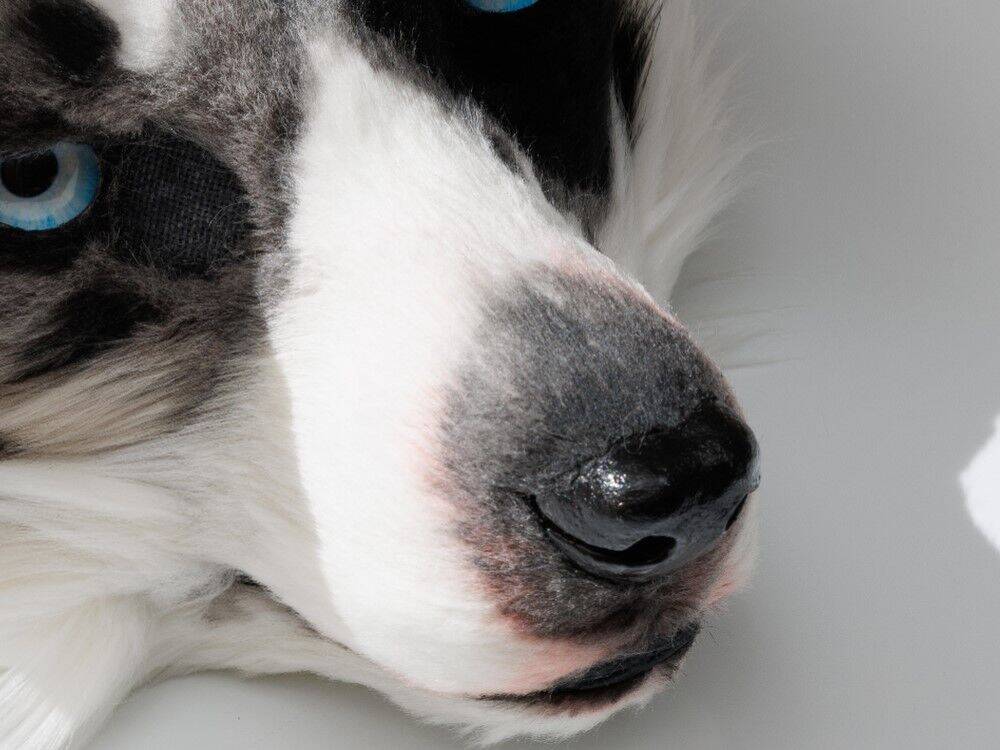} & 
         \includegraphics[width=\scale\linewidth]{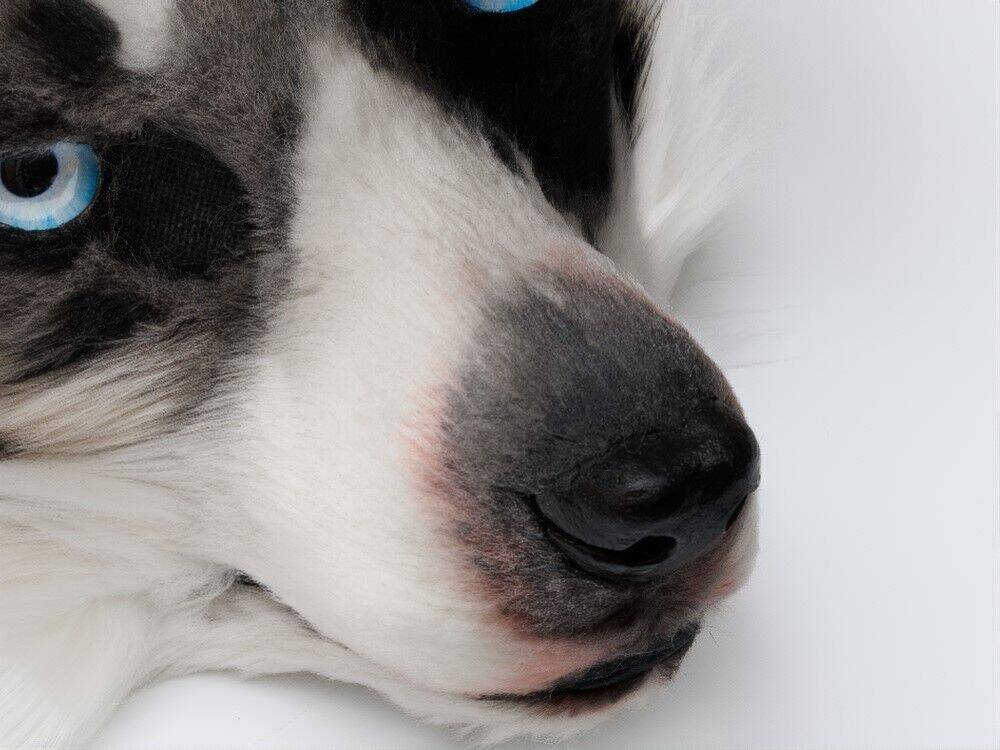} & 
         \includegraphics[width=\scale\linewidth]{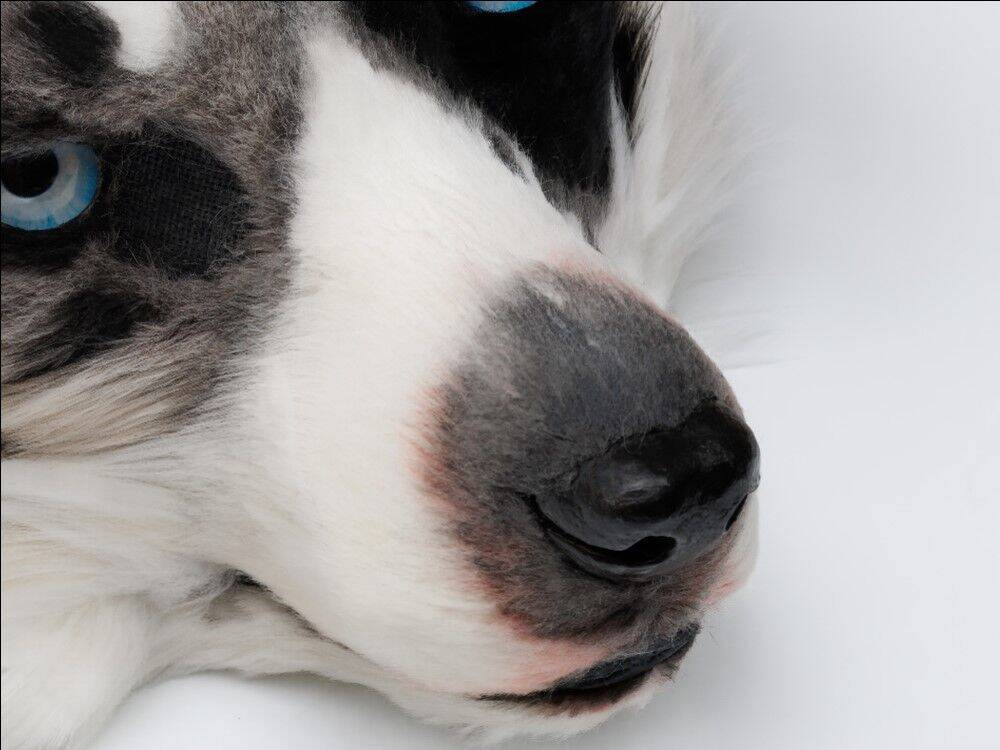} & 
         \includegraphics[width=\scale\linewidth]{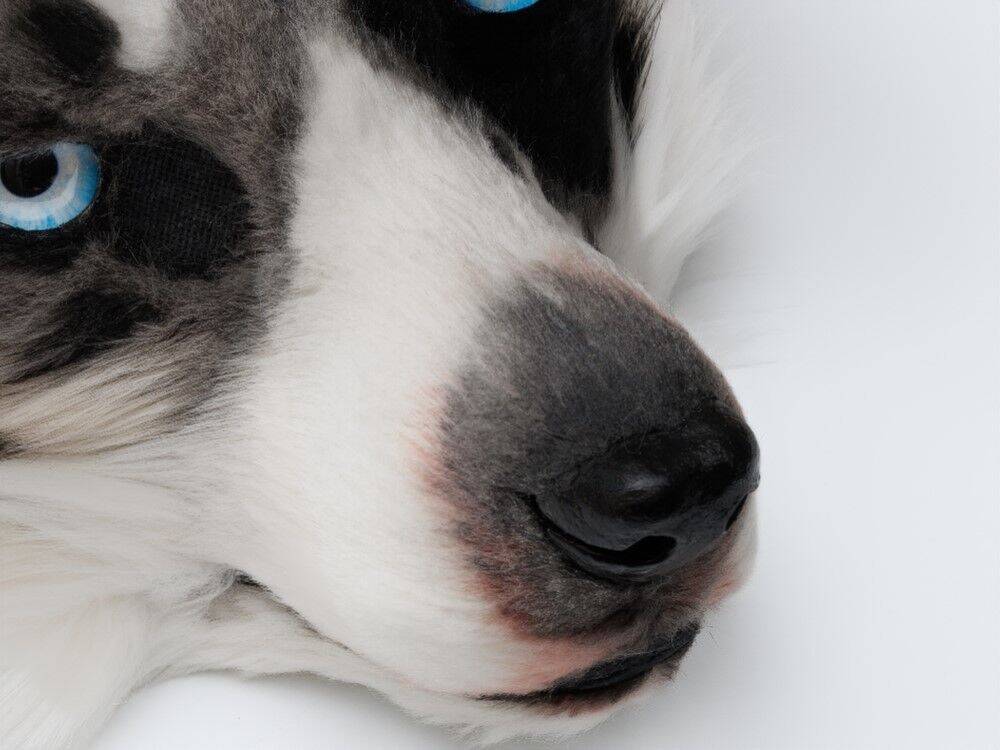} & 
         \includegraphics[width=\scale\linewidth]{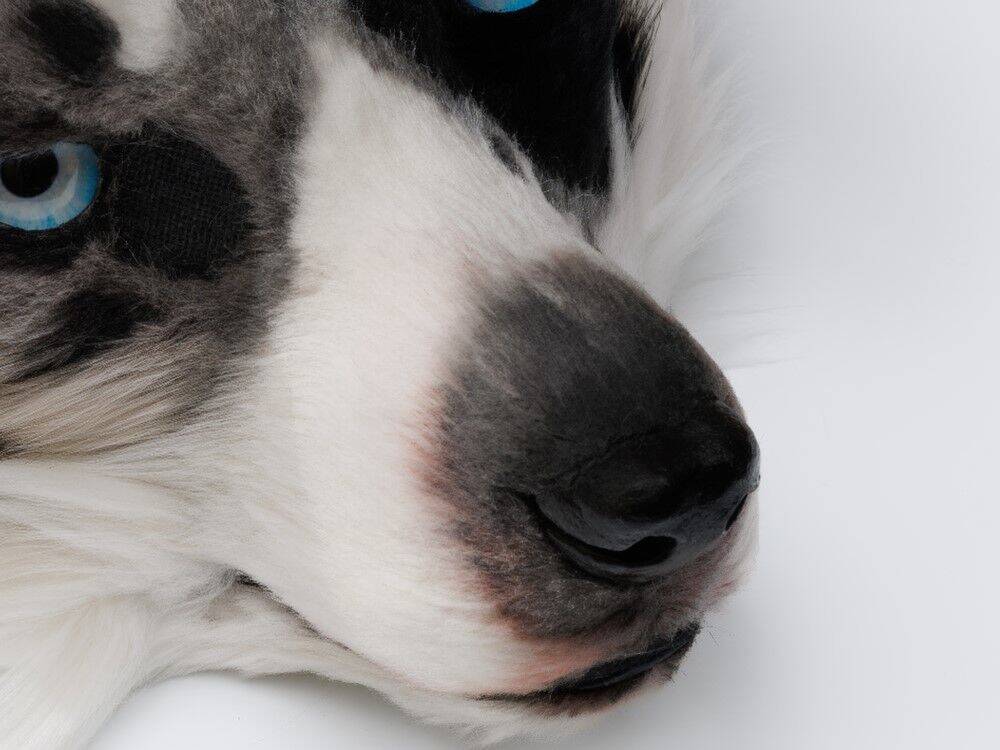} & 
         \includegraphics[width=\scale\linewidth]{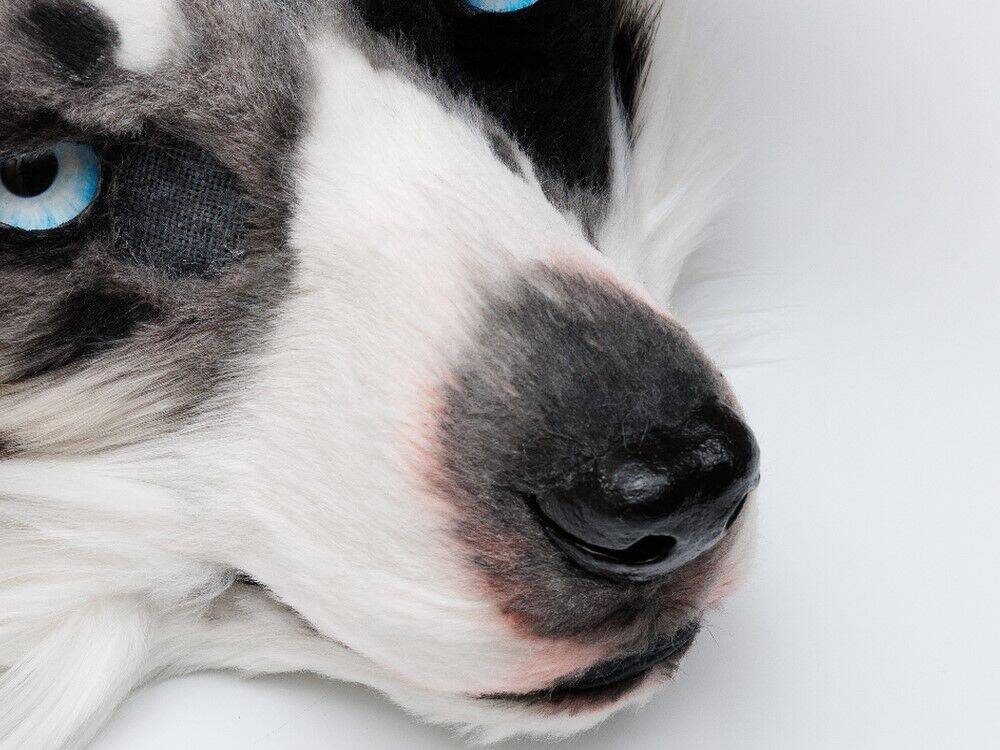} \\
    \end{tabular}
    \vspace{-2mm}
    \caption{
    Equivalent samples from the NTIRE 2025 Image Shadow Removal Challenge test split, for Team X-Shadow, the winner of both tracks vs. the solutions ranked second, third and fourth.
    }
    \vspace{-5mm}
    \label{fig:results-t1}
\end{figure*}

\tref{tab:quanti_t1} provides the quantitative evaluation of the submitted results for both fidelity and perceptual tracks. Particular rankings along each metric evaluated are provided as subscripts. The solutions achieve a significant performance level, in terms of reconstruction fidelity and quantified perceptual quality, with the provided evaluations being backed by the results provided for visual comparison (see \fref{fig:results-t1}).  


\begin{table*}[t!]
    \centering
    \footnotesize
    \begin{tabular}{l|ccc|c|ccc|c|c|c}
        Team & \makecell{PSNR \\ $\uparrow$} & \makecell{SSIM \\ $\uparrow$} & \makecell{LPIPS \\ $\downarrow$} & \makecell{MOS \\ $\uparrow$} & \makecell{Params. \\ (M)} & \makecell{Runtime \\ (s)} & Device & Extra Data & \makecell{Fidelity \\ Rank} & \makecell{Perceptual \\ Rank}  \\
        \hline
       
        X-Shadow & 25.34 & 0.839 & 0.078 & 8.75 & 235 & 1 & A100/RTX4090 & \cite{vasluianu2024towards} & 1 & 1 \\
        LUMOS & 25.38 & 0.838 & 0.095 & 8.20 & 23 & 1.2 & A100 & \cite{vasluianu2023shadow} & 2 & 2 \\
        ACVLab & 25.90 & 0.842 & 0.124 & 8.10 & 25 & 1.63 & RTX3090 & No & 3 & 3 \\
        FusionShadowRemoval & 24.81 & 0.825 & 0.090 & 8.00 & 373 & 0.76 & RTX3090 & \cite{Vasluianu_2024_CVPR} & 5 & 4 \\
        GLHF & 25.31 & 0.834 & 0.096 & 7.98 & 379 & 1.54 & L20 & \cite{vasluianu2023ntire, Vasluianu_2024_CVPR} & 4 & 5 \\
        LVGroup\_HFUT & 25.02 & 0.833 & 0.109 & 7.90 & 17 & 3.46 & RTX4090 & No & 6 & 6 \\
        MIDAS & 24.44 & 0.819 & 0.130 & 7.85 & 294 & 1.56 & RTX3090 & No & 10 & 7 \\
        Alchemist & 23.96 & 0.817 & 0.088 & 7.78 & 281 & 0.3 & RTX4060ti & No & 8 & 8 \\
        PSU Team & 24.15 & 0.811 & 0.133 & 7.73 & 41 & 6 & A100 & No & 12 & 9 \\
        Oath & 24.64 & 0.829 & 0.106 & 7.68 & 17.5 & 0.63 & RTX3090 & No & 7 & 10 \\
        KLETech-CEVI & 23.46 & 0.815 & 0.127 & 7.60 & 10 & 0.7 & RTX6000 & No & 11 & 11 \\
        ReLIT & 23.44 & 0.822 & 0.113 & 7.50 & 175 & 1.24 & RTX4090 & No & 9 & 12 \\
        MRT-ShadowR & 22.15 & 0.780 & 0.118 & 6.80 & 2.39 & 0.8 & RTX3090 & No & 13 & 13 \\
        CV\_SVNIT & 23.87 & 0.772 & 0.216 & 6.60 & 4.7 & 0.4 & RTX5000 & No & 14 & 14 \\
        X-L & 21.45 & 0.781 & 0.149 & 6.03 & 5 & 0.5 & A100 & No & 15 & 15 \\
        ZhouBoda & 14.86 & 0.555 & 0.407 & 2.30 & 67 & 0.22 & vGPU-32GB & \cite{le2021physics, ISTDwang2018STCGAN, vasluianu2023shadow} & 16 & 16 \\
        GroupNo9 & 17.04 & 0.550 & 0.660 & 0.80 & - & 0.09 & T4 *2 & No & 17 & 17 \\        
        \hline
    \end{tabular}
    
    \caption{
    Quantitative evaluations for the submissions of the NTIRE 2025 Image Shadow Removal Challenge on the Final Phase test split. }
    \label{tab:quanti_t1}
\end{table*}

\section{Challenge Methods}
\label{sec:methods}


\begin{figure}
    \centering
    \includegraphics[width=0.99\linewidth]{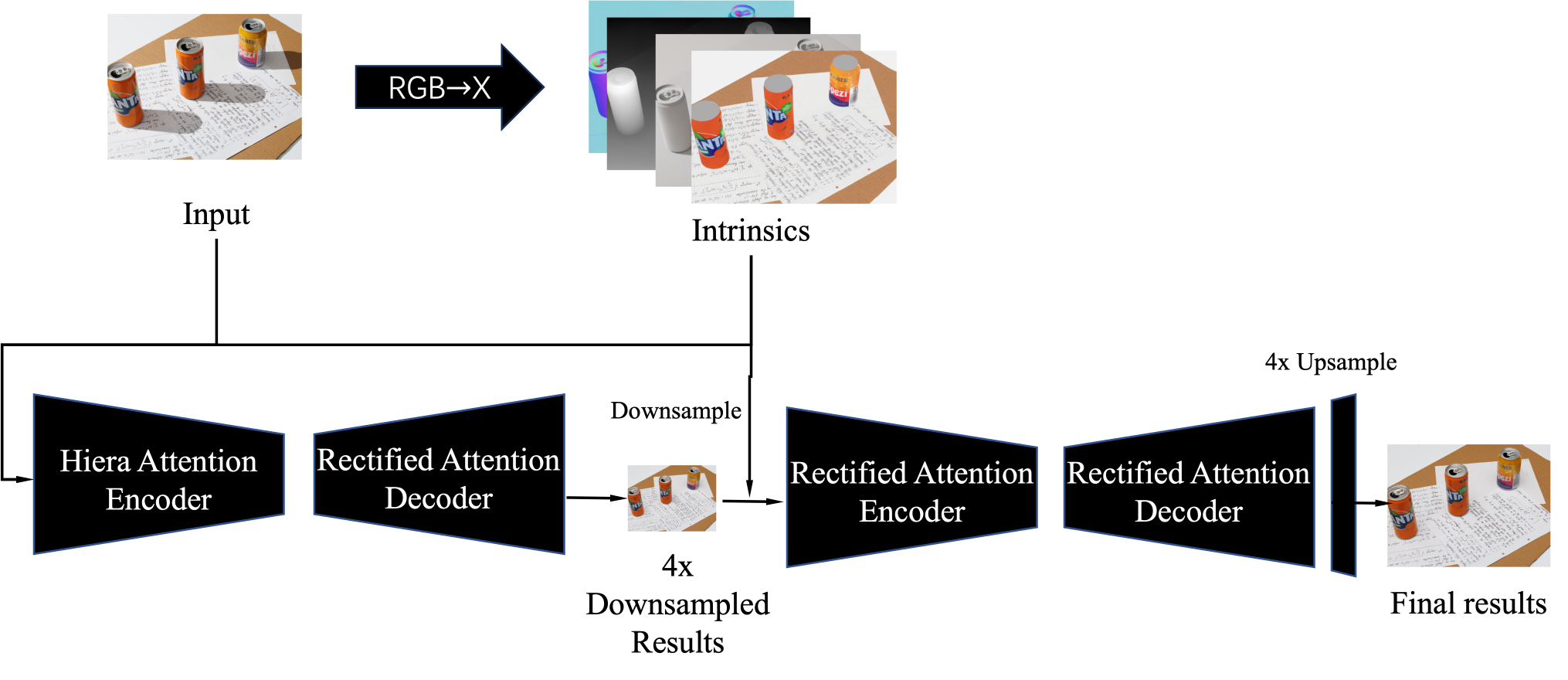}
    \vspace{-3mm}
    \caption{The overall architecture proposed by X-Shadow. Intrinsic hints are generated with auxiliary methods~\cite{DBLP:conf/siggraph/0005DGHHLYH24,ravi2024sam2} before being used as additional inputs for both the initial down-scaling stage and the final up-scaling stage.}
    \vspace{-3mm}
    \label{fig:xshadow}
\end{figure}

\subsection{X-Shadow}
The solution of team X-Shadow leverages intrinsic hints such as albedo, irradiance, normal, and depth information that has been extracted from the input image using auxiliary methods.
These are produced by \emph{rgb2x}~\cite{DBLP:conf/siggraph/0005DGHHLYH24} and DepthAnythingV2~\cite{ravi2024sam2} methods.

As observed in \fref{fig:xshadow}, their solution consists of two stages. 
Firstly, a down-sampling stage processes the input image in a low-resolution space, with a high-level semantic pre-trained attention-based u-shape network, which is implemented through a Hierarchical Transformer Block from SegmentAnythingV2~\cite{ravi2024sam2}.

Then, an up-sampling stage focuses on detail reconstruction using a lightweight attention-based u-shape network, implemented via LRNet, in a fashion similar to ShadowHack~\cite{DBLP:journals/corr/abs-2412-02545}. 
Note that for both stages, the previously generated partial results are used as additional inputs.

\begin{figure}
    \centering
    \includegraphics[width=0.99\linewidth]{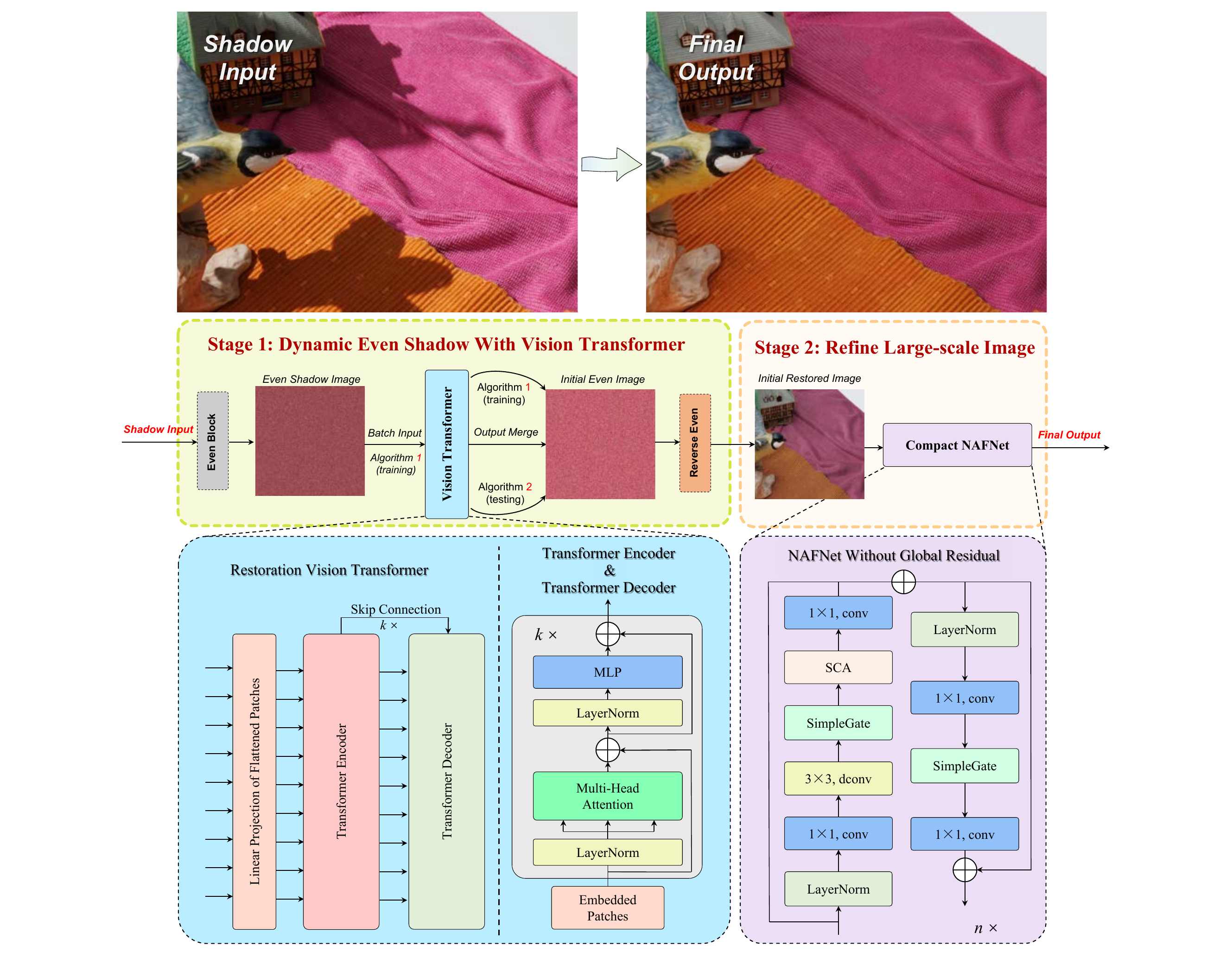}
    \vspace{-2mm}
    \caption{Overview of the Team LUMOS proposed EvenFormer.}
    \label{fig:LUMOS}
    \vspace{-3mm}
\end{figure}

\subsection{LUMOS}
Team LUMOS proposes the EvenFormer architecture to perform image restoration for the unevenly distributed Shadow Removal task. An overview of their solution is available in \fref{fig:LUMOS}.
To take advantage of the strong contextual information extraction capability of transformers in modeling long sequences, they utilize a Vision Transformer framework consistent with~\cite{he2021masked}.
As the images in the challenge are of high-resolution, they propose a dynamic approach to assist the transformer module in efficiently handling large, 2K-sized, images.
LUMOS found that a single-stage network results in poor performance when handling strong shadows, in particular artifacts caused by the transformer block processing. 
To remedy this issue they introduce a NAFNet~\cite{chen2022simple} based refinement stage, implementing a Coarse-to-Fine image restoration, similar to~\cite{EnhancedCF}.
\subsection{ACVLab}

The solution by ACVLab is based on the OmniSR~\cite{xu2024omnisr} architecture.
To further enhance the results of the baseline, ACVLab incorporated the Frequency-Aware Feature Fusion method FreqFusion~\cite{chen2024frequency} into its decoder stage.
As shadow areas typically retain texture information, which is primarily present in the high frequency band of the image spectrum, this addition enables the model to effectively preserve details, enhancing model predictions quality.
\begin{figure}
    \centering
    \includegraphics[width=0.99\linewidth]{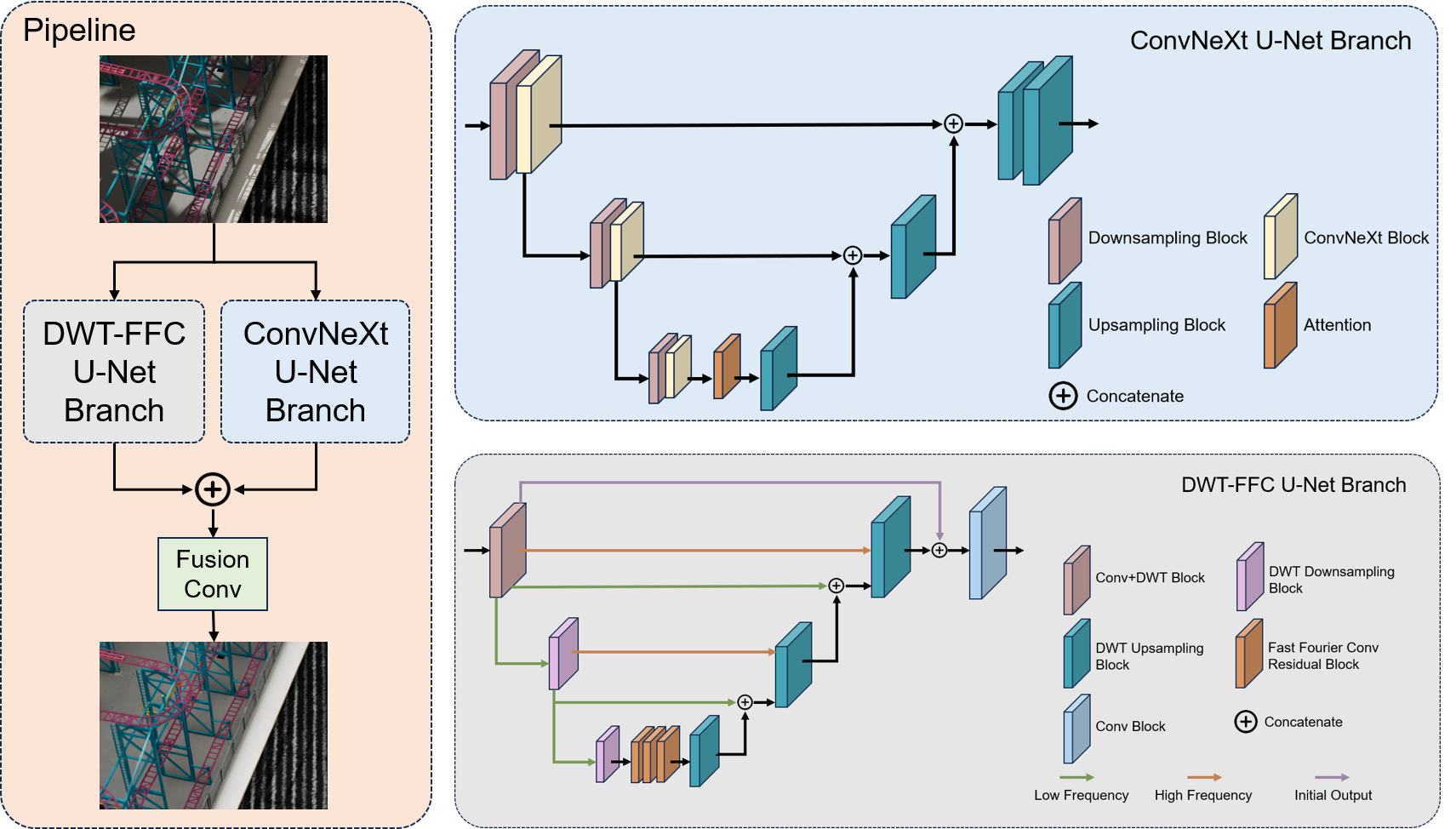}
    \vspace{-2mm}
    \caption{The Fusion Network for Image Shadow Removal proposed by team FusionShadowRemoval.}
    \vspace{-3mm}
    \label{fig:fusionshadowremoval}
\end{figure}

\subsection{FusionShadowRemoval}

Team FusionShadowRemoval proposes a dual-branch architecture, combining a ConvNext~\cite{liu2022convnext} U-Net branch and a DWT-FFC U-Net branch, which has previously shown promising results for Image Dehazing and Shadow Removal tasks~\cite{wei2024shadow, zhou2023dwt-ffc, hybrid_loss1}. 
They adopt block designs proposed by ShadowRefiner~\cite{wei2024shadow}, but notably change the arrangement, completely disregarding the second refinement stage. 
As shown in \fref{fig:fusionshadowremoval}, the architecture can be separated into three operations.

The \textbf{ConvNext U-Net Branch} consists of multiscale ConvNext~\cite{liu2022convnext} and Attention blocks~\cite{wei2024shadow}. 
Specifically, three down-sampling layers and four up-sampling layers are used, with each down-sampling followed by a ConvNext block for feature extraction. 
In the deepest layer, the features are processed with an attention block before the encoded features are restored to their original resolution via up-sampling blocks supported by skip connections to pass global semantic information.

The \textbf{DWT-FFC U-Net Branch} uses Discrete Wavelet Transform (DWT) to extract high-frequency and low-frequency features. 
The low-frequency representation is spliced with the normal convolutional output, while the high-frequency features are passed to the up-sampling module to gradually remove shadows. 
To make the reconstructed image more realistic and perceptual, a Fast Fourier Convolution Residual block is utilized to remove shadows using spatial and spectral information. 

The final \textbf{Fusion Stage} uses a single convolution to fuse the features of the ConvNext U-Net branch and the DWT-FFC U-Net branch, obtaining the shadow-free output.
\begin{figure}
    \centering
    \includegraphics[width=0.99\linewidth]{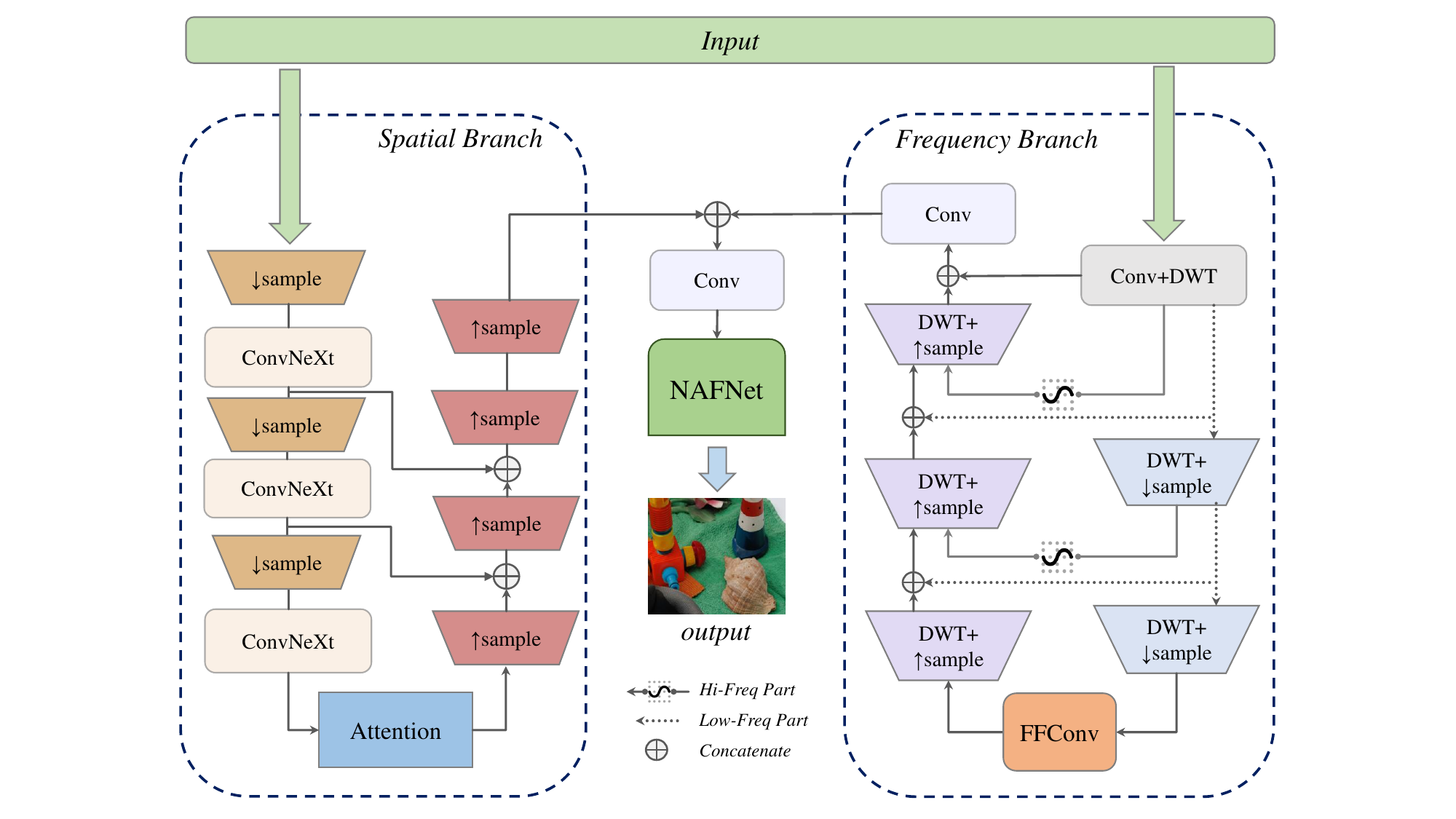}
    \caption{The ShadowAlpha architecture proposed by GLHF uses two parallel processing branches with a final fusion and refinement stage.}
    \label{fig:GLHF}
\end{figure}

\subsection{GLHF}
Team GLHF proposes the \emph{ShadowAlpha} method based on spatial and frequency convolution networks.
Inspired by the architecture ShadowRefiner~\cite{wei2024shadow}, they adopt it into a two-branch design (as shown in~\fref{fig:GLHF}).
The first branch extracts robust spatial features via three ConvNeXt stages ~\cite{liu2022convnext}.
To improve the global consistency in the final output image, a second parallel branch based on a DWT-FFC~\cite{zhou2023dwt-ffc} network is used to capture frequency characteristics of the shadow-affected regions.
The outputs of both branches are then fused by a simple convolution before the NAFNet~\cite{chen2022simple} refinement stage.

\begin{figure}
    \centering
    \includegraphics[width=0.99\linewidth]{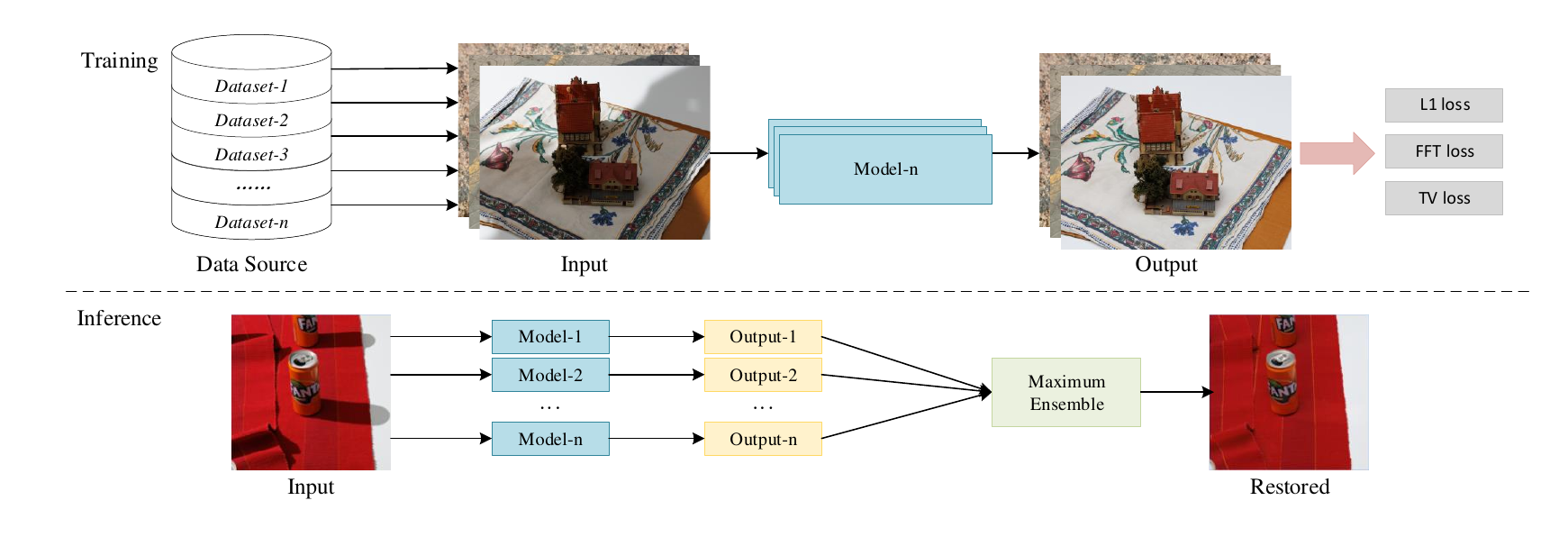}
    \vspace{-3mm}
    \caption{The multi-dataset maximum ensamble architecture proposed by the LVGroup\_HFUT team.}
    \vspace{-3mm}
    \label{fig:LVGroup}
\end{figure}

\subsection{LVGroup\_HFUT}

The submission from LVGroup\_HFUT combines the popular NAFNet~\cite{chen2022simple} architecture with a maximum ensemble approach.
As depicted in \fref{fig:LVGroup}, they utilize a multi-dataset training strategy to improve performance and robustness in diverse scenarios.
Here, separate NAFNet models \cite{chen2022simple} are initially optimized on different training datasets.
During the inference process, each model processes the input image separately in parallel.
The final output image is obtained by applying a maximum ensemble strategy~\cite{Vasluianu_2024_CVPR} to combine the output of all models into a single image.
To implement their solution, they train three models on the official challenge training data, and three additional models on a combination of WSRD+ \cite{Vasluianu_2024_CVPR}, ISTD+~\cite{ISTDwang2018STCGAN} and SRD~\cite{SRDDESHADOW}.
As each of the six models offers unique advantages and characteristics, this enables the overall architecture to represent complementary patterns and features, leading to better generalization ability on unseen data.

\begin{figure}
    \centering
    \includegraphics[width=0.99\linewidth]{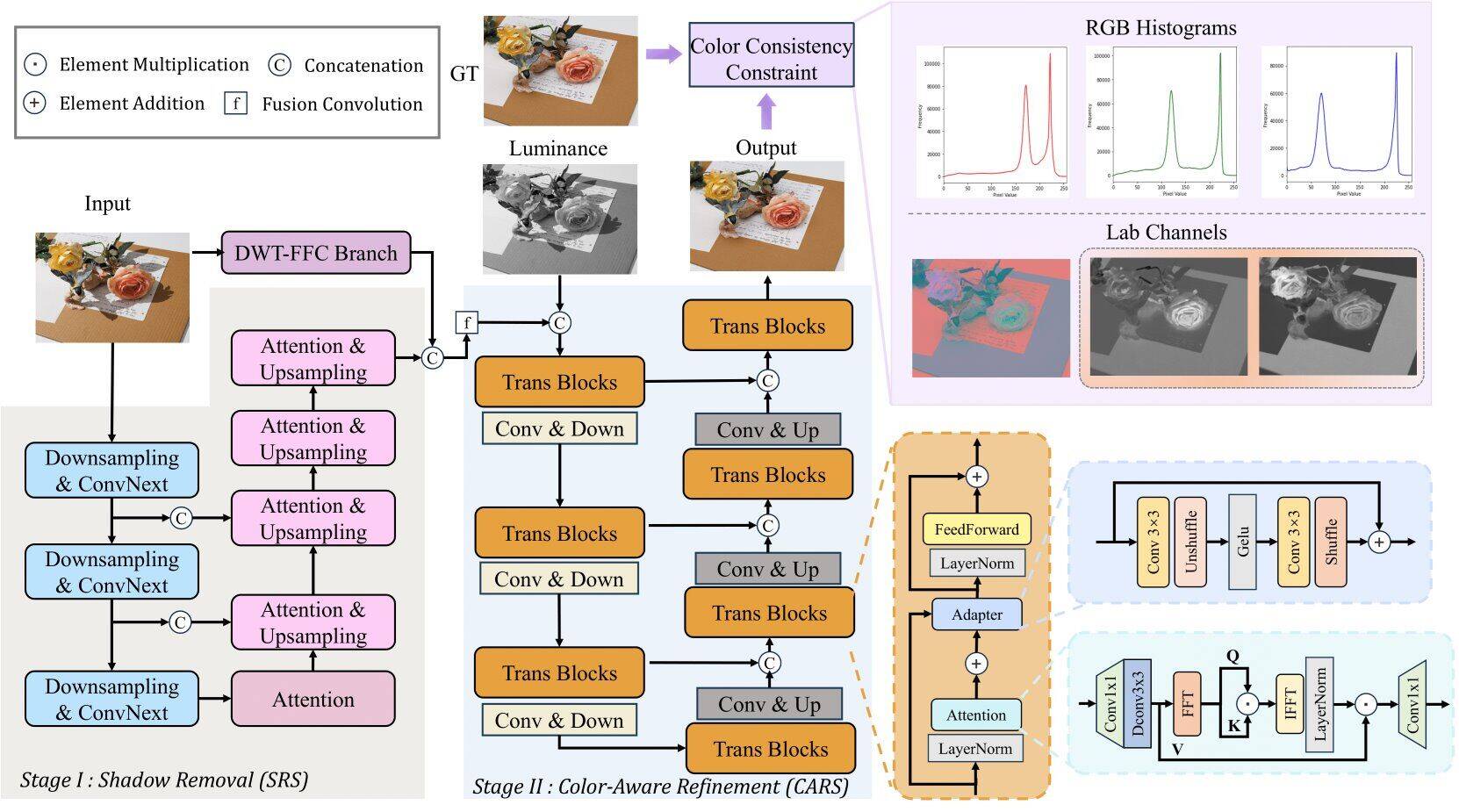}
    \caption{The two stage L-Adapter architecture proposed by MIDAS, the second stage utilizes  an auxiliary luminance information.}
    \label{fig:MIDAS}
\end{figure}

\subsection{MIDAS}
The MIDAS team upgrades the Shadow Refiner~\cite{wei2024shadow} architecture by injecting luminance information into the new color-aware refinement stage (CARS) and applying a Color Consistency Constraint on the output image (see \fref{fig:MIDAS}).
The Shadow Removal Stage (SRS) adopted from Shadow Refiner~\cite{wei2024shadow} is designed to eliminate shadows from the input by leveraging spatial and frequency representation learning.
Although SRS effectively eliminates shadows, experiments performed by the team indicated that it also alters object brightness, leading to unintended color shifts within the scene.
To address this issue, the new CARS with additions color constraint optimization are introduced; these additions account for the influence of light sources that alter the color of objects within the scene.

Specifically, CARS implements a Restormer~\cite{zamir2022restormer} architecture with additional adapters to inject luminance information.
To further reinforce color consistency, constraints are applied in both the RGB and the Lab color spaces. 
In the RGB space, the histogram constraint is applied for three channels separately, while in the $Lab$ space, constraints on the $a$ and $b$ channels ensure accurate color preservation.

\begin{figure}
    \centering
    \includegraphics[width=0.99\linewidth]{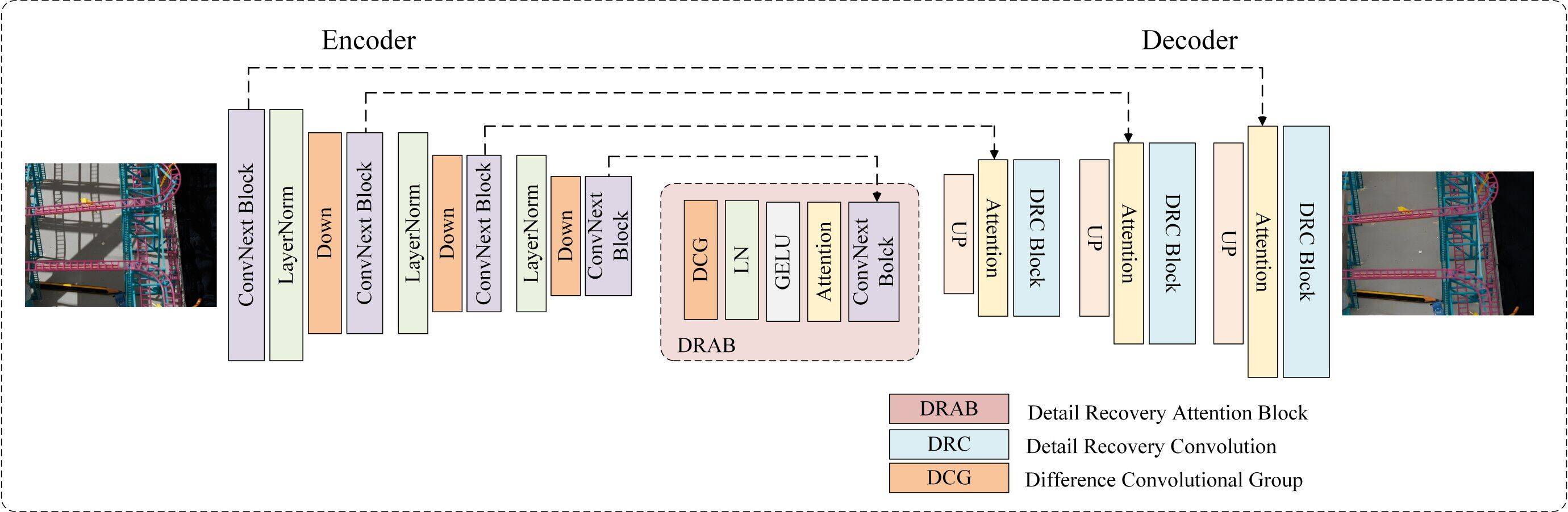}
    \vspace{-3mm}
    \caption{A graphical representation of the solution proposed by Team Alchemist.}
    \label{fig:Alchemist}
    \vspace{-3mm}
\end{figure}

\subsection{Alchemist}
As shown in \fref{fig:Alchemist}, the proposed solution by team Alchemist is a U-Net~\cite{ronneberger2015u} shaped encoder/decoder architecture.
It primarily employs a ImageNet~\cite{deng2009imagenet} pre-trained ConvNext~\cite{liu2022convnext} model as a powerful encoder in conjunction with a specialized  decoder focusing on restoring boundary details, thereby achieving high-quality shadow removal.  
Notably, the pre-trained ConvNext encoder generates robust latent spatial information, enabling the reconstruction of shadow boundary details through detail recovery convolution during decoding. 
\begin{figure}
    \centering
    \includegraphics[width=0.99\linewidth]{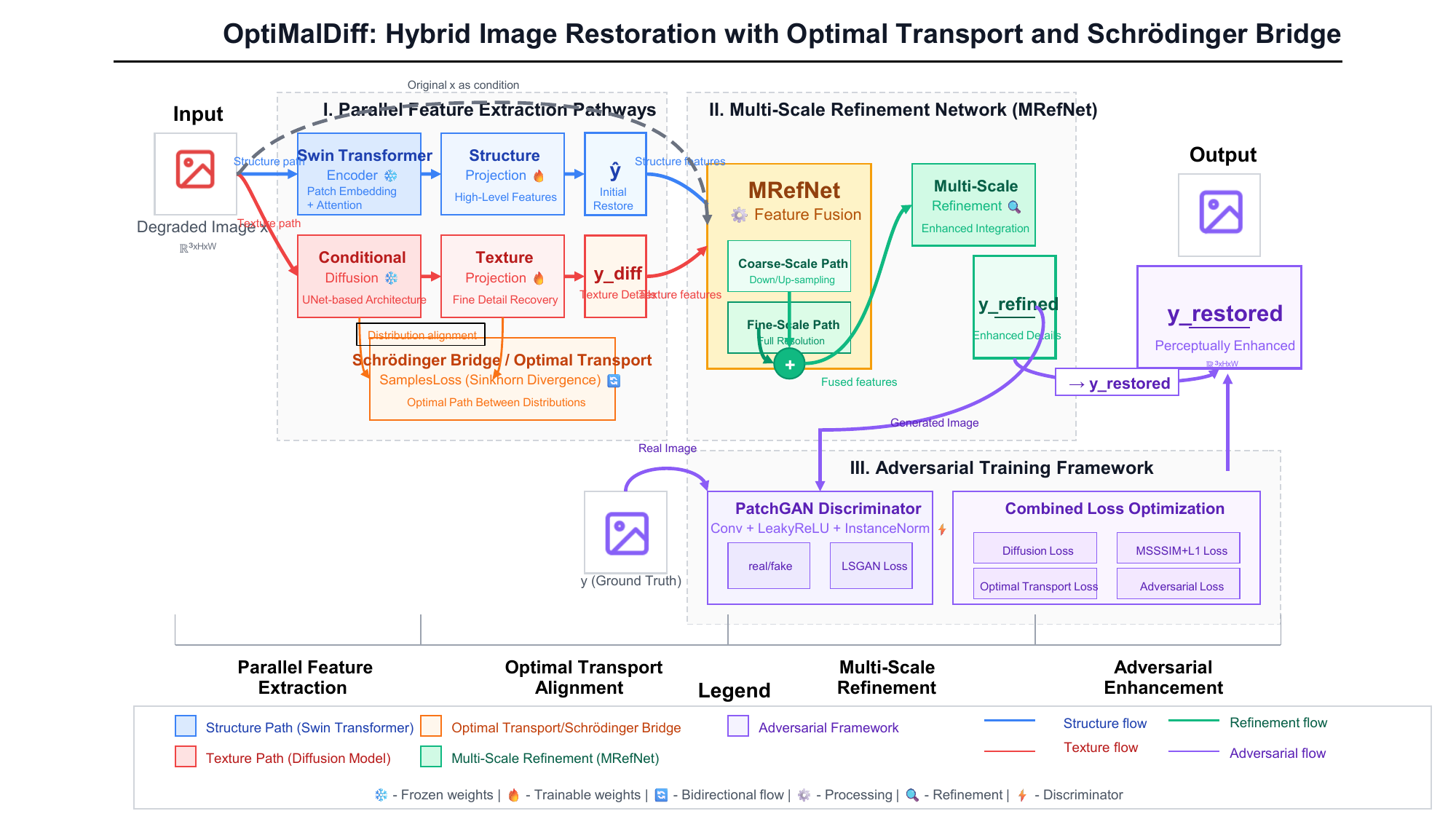}
    \caption{Overview of the OptiMalDiff architecture combining Schrödinger Bridge diffusion, transformer-based feature extraction, and adversarial refinement.}
    \label{fig:PSUTeam}
\end{figure}

\subsection{PSU Team}



PSUTeam proposes \textbf{OptiMalDiff}, a high-fidelity image enhancement framework that reformulates image denoising as an optimal transport problem. 
The core idea is to model the transition from noisy to clean image distributions via a Schrödinger Bridge-based~\cite{de2021diffusion} diffusion process. 
The architecture (shown in ~\fref{fig:PSUTeam}) consists of three main components: 
(1) a hierarchical Swin Transformer~\cite{liu2021Swin} backbone that extracts both local and global features efficiently, 
(2) a Schrödinger Bridge Diffusion Module that learns forward and reverse stochastic mappings, and 
(3) a Multi-Scale Refinement Network (MRefNet) designed to progressively refine image details. 
To enhance realism, a PatchGAN discriminator~\cite{isola2017image} is integrated for adversarial training.

The resulting network is trained end-to-end using a composite loss function that includes diffusion loss, Sinkhorn divergence optimal transport loss~\cite{di2020optimal}, multi-scale SSIM + L1 loss for perceptual quality, and adversarial loss.
\begin{figure}
    \centering
    \includegraphics[width=0.99\linewidth]{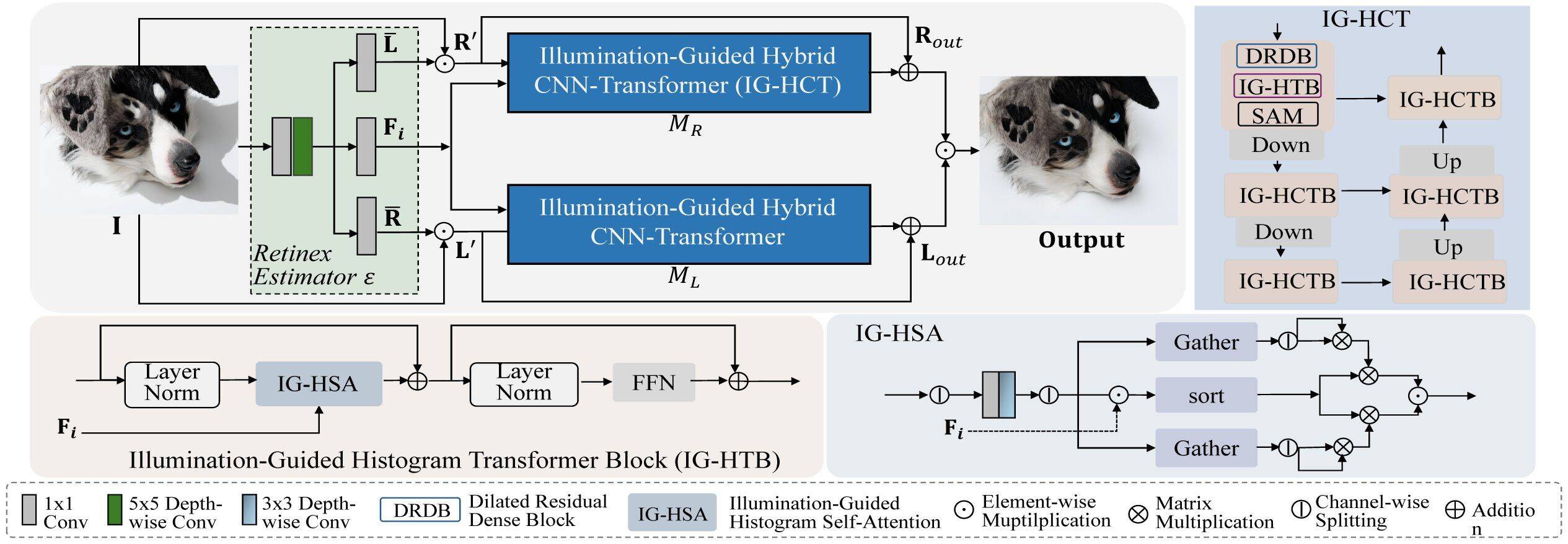}
    \vspace{-2mm}
    \caption{Overview of the dual branch architecture proposed by Oath. One branch restores reflectance while the other restores illumination information.
    }
    \label{fig:Oath}
    \vspace{-3mm}
\end{figure}

\subsection{Oath}
Team Oath proposes ReHiT, an efficient retinex-guided histogram transformer for shadow removal, where Retinex theory~\cite{land1977retinex} is employed to develop the dual-branch restoration pipeline. 
As shown in the proposed architecture (\fref{fig:Oath}), after separating reflectance and illuminance via the Retinex estimator, each is individually restored by one network branch~\cite{ecmamba24neurips}.
Each branch is a Illumination-Guided Hybrid CNN-Transformers (IG-HCT) module composed of multiple IG-HCT blocks each containing a Dilated Residual Dense Block (DRDB)~\cite{ESDNet22eccv}, an Illumination-Guided Histogram Transformer Block (IG-HTB), and a Semantic-aligned scale-aware module (SAM)~\cite{ESDNet22eccv} to boost the network’s capability in handling shadow patterns with diverse scales without incurring too much computational cost. 
The design of IG-HTB is inspired by RetinexFormer~\cite{cai2023retinexformer}, Histoformer~\cite{histoformer24eccv}, and GPP-LLIE~\cite{gpp25aaai}. 
This block consists of two-layer normalization (LN), an Illumination-Guided Histogram Self-Attention (IG-HSA) layer, and a feed-forward network (FFN). Besides the L1 and multi-scale SSIM loss~\cite{glare24eccv}, the structure loss~\cite{lita25cvpr} and additional constraints~\cite{ecmamba24neurips}  are used for optimization supervision.


\subsection{KLETech-CEVI}
Team KLETech-CEVI proposes \emph{WATFormer: Wavelet Attention-Based Transformer for Shadow Removal}, a two-stage shadow removal and refinement model.
Inspired by \cite{wei2024shadow}, an attention mechanism based on wavelet transforms is proposed, the image signal being decomposed into various frequencies, aiming to improve feature-characteristic understanding and analysis, specifically in terms of shadow image segments identification. 
The shadow removal module is based on a ConvNext-driven U-Net, incorporating 7×7 depth-wise convolution at each resolution level \cite{liu2022convnext}. 
In addition, it integrates the frequency branch network from \cite{zhou2023dwt-ffc}. 
This enables the Shadow Removal module to effectively utilize both spectral and spatial information for mapping shadow-affected and shadow-free images. 
T
In the refinement module, both the encoder and decoder utilize Wavelet Attention-Based Transformer (WAT) blocks, decomposing the input into multiple frequencies that aid in better understanding and feature extraction. 

\subsection{ReLIT}
Team ReLIT proposes a novel shadow removal framework, MatteViT, that integrates shadow matte guidance with advanced neural architectures to effectively address the challenging task of shadow removal. The overall architecture consists of three main components: Shadow Matte Generator, Matte-Guided Vision Transformer, and Spatial NAFNet for refinement.

In the first stage, TransUNet~\cite{chen2024transunet} is used to predict a shadow matte from the input shadow image as a guide for the subsequent shadow removal process. 
Training data for this is obtained from paired shadow and shadow-free images, leveraging luminance differences in the LAB color space. 
This explicit shadow modeling helps to understand shadow characteristics such as shape, boundary, and density, which are essential for accurate shadow removal.

The Matte-Guided Vision Transformer (MatteViT) sits at the core of this approach, it processes the input image while incorporating the shadow matte information generated in the previous stage.
This is achieved through first incorporating shadow matte information into the multi-head attention mechanism by adjusting attention weights, therefore enabling the network to focus more effectively on shadow-affected regions during processing.
Secondly, by implementing a Feature-wise Linear Modulation (FiLM) mechanism~\cite{perez2018film} that adaptively adjusts feature representations based on frequency information extracted from the input image. 
This is achieved by computing the Fast Fourier Transform (FFT) of the input image, extracting magnitude features, and using these to generate modulation parameters (gamma and beta) for conditional processing.

After processing by MatteViT, the spatial NAFNet~\cite{chen2022simple} for refinement then enhances visual quality by addressing potential artifacts and improving texture consistency.

The model is optimized for spatial accuracy and frequency consistency by using Charbonnier loss for spatial domain and FFT loss for frequency domain, preserving local details and global illumination characteristics.



\subsection{MRT-ShadowR}
MRT-ShadowR proposes the Multi-stage Residual Transformer (MRT) network for shadow removal.
They adopt a Multi-scale Entanglement Scheme described in ~\cite{ntire2024lowlight} (Sec. 4.16), but tailored for Transformers, helping the network learn better multi-scale feature representations.
To aid in detail preservation the Residual Multi-headed Self-Attention, a residually-enhanced attention mechanism is introduced to help preserve details across various stages in the network.
Lastly, for enhanced performance and effective feature extraction, the Multi-stage Squeeze \& Excite Fusion Block~\cite{brateanu2024} is included as a post-attention step.

To optimize our network, a hybrid loss capable of capturing pixel-level, multi-scale and perceptual cues, as proposed in \cite{kant2024isetc, kant2025sensors}, is adopted, leading to increased effectiveness in high-resolution, high-fidelity restoration.

\subsection{CV\_SVNIT}
Team CV\_SVNIT a introduces a two-stage approach for shadow removal and refinement.
The Shadow Removal module effectively eliminates shadows using a ConvNext-based~\cite{liu2022convnext} U-Net architecture with multi-scale convolutional layers (3 × 3, 5 × 5, and 7 × 7 kernels) applied in parallel to capture features at various scales. 
A frequency branch network is integrated to jointly process spectral and spatial information for accurate shadow removal.
The enhancement stage is implemented via a Fast-Fourier Attention Transformer architecture with hierarchical encoding and decoding.


\subsection{X-L}
The method proposed by X-L is inspired by SRFormer~\cite{zhou2023srformer} and HGNet~\cite{10130817}. 
Initially, a hue component estimation is used to highlight shadow-affected segments in the HSV space. 
Then, both the hue features and the shadowed input are processed separately by the encoder and a single permuted self-attention blocks~\cite{zhou2023srformer} (PSAB) to further achieve compact feature representation. 
Then, the features are concatenated and processed by five more PSAB blocks for combined feature enhancement before obtaining the shadow-free image with a decoding module. 
\subsection{ZhouBoda}

Team ZhouBoda adopts DNSR~\cite{vasluianu2023shadow}, a shadow removal method that achieves accurate shadow detection, local restoration, global coordination, and natural color recovery through multiscale feature extraction, dynamic feature modulation, mask-guided fusion, and dual path refinement.
\subsection{Group No.9}
The Group No.9 team proposes a custom CNN-based U-Net architecture enhanced with a pre-trained ResNet-50 for the extraction of shadow-features. 
To improve the performance they implemented self-assembly techniques and post-processing such as noise reduction.

\section{Conclusion}
The NTIRE 2025 Image Shadow Removal challenge builds on the success of the previous editions, enjoying significant attention from the computer vision community. 
Participants jointly optimize for both fidelity and perceptual quality, with two separate evaluation tracks. 
The challenge received a high number of submissions, with the analyzed solutions achieving a significant quantified performance level in both evaluation settings. 
This shows in the quantified PSNR, SSIM and LPIPS characterizing the restored images correlating with the rankings of the conducted user study. 
Many challenge participants provided insightful feedback with future ideas for the next editions of the challenge. 
 
\section*{Acknowledgments}
This work was partially supported by the Humboldt Foundation. We thank the NTIRE 2025 sponsors: ByteDance, Meituan, Kuaishou, and University of Wurzburg (Computer Vision Lab).

{\small
\bibliographystyle{ieee_fullname}
\bibliography{main}
}

\newpage



\appendix
\section{Teams and Affiliations}
\label{sec:apd:team}

\subsection*{NTIRE 2025 Image Shadow Removal Challenge Team}
\noindent\textit{\textbf{Members:}}\\ 
\textit{Florin-Alexandru Vasluianu$^1$}, Tim Seizinger$^1$, 
Zhuyun Zhou$^1$, Cailian Chen$^2$, Zongwei Wu$^1$, Radu Timofte$^1$\\
\noindent\textit{\textbf{Affiliations: }}\\
$^1$ Computer Vision Lab, IFI \& CAIDAS, University of W\"urzburg \\
$^2$ Shanghai Jiaotong University, China\\

\subsection*{X-Shadow}
\noindent\textit{\textbf{Title: }}\\
Leveraging Intrinsic Hint for High-Quality Mask-Free Shadow Removal \\
\noindent\textit{\textbf{Members:}}\\ 
\textit{Mingjia Li$^1$}, Jin Hu$^1$, Hainuo Wang$^1$, Hengxing Liu$^1$, Jiarui Wang$^1$, Qiming Hu$^1$, Xiaojie Guo$^1$\\
\noindent\textit{\textbf{Affiliations: }}\\
$^1$ Tianjin University, Tianjin, China\\

\subsection*{LUMOS}
\noindent\textit{\textbf{Title: }}\\
EvenFormer: Dynamic Even Transformer for High-Resolution Image Shadow Removal \\
\noindent\textit{\textbf{Members:}}\\ 
\textit{Xin Lu$^1$}, Yurui Zhu$^1$, Xingbo Wang$^1$, Yuanfei Bao$^1$, Jiarong Yang$^1$, Anya Hu$^1$, Kunyu Wang$^1$, Zihao Fan$^1$, Jie Xiao$^1$, Dong Li$^1$, Xi Wang$^1$, Hongjian Liu$^1$, Xueyang Fu$^1$, Zheng-Jun Zha$^1$\\
\noindent\textit{\textbf{Affiliations: }}\\
$^1$ University of Science and Technology of China, Hefei, China\\

\subsection*{ACVLab}
\noindent\textit{\textbf{Title: }}\\
Enhanced OmniSR for Shadow Removal with Frequency Fusion
Signal \\
\noindent\textit{\textbf{Members:}}\\ 
\textit{Yu-Fan Lin$^1$}, Chia-Ming Lee$^1$, Chih-Chung Hsu$^{1,2}$\\
\noindent\textit{\ textbf{Affiliations: }}\\
$^1$ Institute of Data Science, National Cheng Kung University\\
$^2$ Institute of Intelligent Systems, National Yang Ming Chiao Tung University\\

\subsection*{FusionShadowRemoval}
\noindent\textit{\textbf{Title: }}\\
A Fusion Network for Image Shadow Removal \\
\noindent\textit{\textbf{Members:}}\\ 
\textit{Yuxu Chen$^1$}, Bin Chen$^1$, Hongwei Wang$^1$, Yuanbin Chen$^1$, Yuanbo Zhou$^1$, Tong Tong$^1$\\
\noindent\textit{\textbf{Affiliations: }}\\
$^1$ Fuzhou University, China\\

\subsection*{GLHF}
\noindent\textit{\textbf{Title: }}\\
ShadowAlpha: Shadow Removal with Spatial and Frequency Convolution Network \\
\noindent\textit{\textbf{Members:}}\\ 
\textit{Jiarong Yang$^1$}, Yuanfei Bao$^1$, Anya Hu$^1$, Xin Lu$^1$, Zihao Fan$^1$, Xingbo Wang$^1$, Kunyu Wang$^1$, Jie Xiao$^1$, Dong Li$^1$, Yurui Zhu$^1$, Xi Wang$^1$, Yupeng Xiao$^1$, Xueyang Fu$^1$, Zheng-Jun Zha$^1$\\
\noindent\textit{\textbf{Affiliations: }}\\
$^1$ University of Science and Technology of China, Hefei, China\\

\subsection*{LVGroup\_HFUT}
\noindent\textit{\textbf{Title: }}\\
Enhancing Shadow Removal via Multi-Dataset Ensemble Learning \\
\noindent\textit{\textbf{Members:}}\\ 
\textit{Zhao Zhang$^1$}, Suiyi Zhao$^1$, Bo Wang$^1$, Yan Luo$^1$, Mingshen Wang$^1$, Yilin Zhang$^1$, Yanyan Wei$^1$\\
\noindent\textit{\textbf{Affiliations: }}\\
$^1$ Hefei University of Technology, China\\

\subsection*{MIDAS}
\noindent\textit{\textbf{Title: }}\\
L-Adapter: Adapters injects luminance knowledge into pre-trained model with color constraint \\
\noindent\textit{\textbf{Members:}}\\ 
\textit{Ziyu Feng$^1$}, Shidi Chen$^1$, Bowen Luan$^1$, Zewen Chen$^1$\\
\noindent\textit{\textbf{Affiliations: }}\\
$^1$ Beijing Jiaotong University, Beijing, China\\

\subsection*{Alchemist}
\noindent\textit{\textbf{Title: }}\\
Prior guided differential information compensated image shadow removal network \\
\noindent\textit{\textbf{Members:}}\\ 
\textit{Haobo Liang$^1$}, Yan Yang$^1$, Jiajie Jing$^1$, Junyu Li$^1$\\
\noindent\textit{\textbf{Affiliations: }}\\
$^1$ School of Electronics and Information Engineering, Lanzhou Jiaotong University, China\\

\subsection*{PSU Team}
\noindent\textit{\textbf{Title: }}\\
OptimalDiff: High-Fidelity Image Enhancement Using Schrödinger Bridge Diffusion and Multi-Scale Adversarial Refinement\\
\noindent\textit{\textbf{Members:}}\\ 
\textit{Bilel Benjdira$^1$}, Anas M. Ali$^1$, Wadii Boulila$^1$\\
\noindent\textit{\textbf{Affiliations: }}\\
$^1$ Robotics and Internet-of-Things Laboratory, Prince Sultan University, Riyadh 12435, Saudi Arabia\\

\subsection*{Oath}
\noindent\textit{\textbf{Title: }}\\
ReHiT: Retinex-guided Histogram Transformer for Mask-free Shadow Removal\\
\noindent\textit{\textbf{Members:}}\\ 
\textit{Wei Dong$^1$}, Han Zhou$^1$, Seyed Amirreza Mousavi$^1$, Jun Chen$^1$\\
\noindent\textit{\textbf{Affiliations: }}\\
$^1$ McMaster University\\

\subsection*{KLETech-CEVI}
\noindent\textit{\textbf{Title: }}\\
WATFormer: Wavelet Attention-Based Transformer for Shadow Removal\\
\noindent\textit{\textbf{Members:}}\\ 
\textit{Vijayalaxmi Ashok Aralikatti$^{1,3}$}, G Gyaneshwar Rao$^{2,3}$, Nikhil Akalwadi$^{1,3}$, Chaitra Desai$^{1,3}$, Ramesh Ashok Tabib$^{2,3}$, Uma Mudenagudi$^{2,3}$\\
\noindent\textit{\textbf{Affiliations: }}\\
$^{1}$ School of Computer Science and Engineering, KLE Technological University\\
 $^{2}$ School of Electronics and Communication Engineering, KLE Technological University\\
 $^{3}$ Center of Excellence in Visual Intelligence (CEVI) KLE Technological University\\

\subsection*{ReLIT}
\noindent\textit{\textbf{Title: }}\\
MatteViT: Shadow-Aware Transformer for High-Fidelity Shadow Removal\\
\noindent\textit{\textbf{Members:}}\\ 
\textit{Seoyeon Lee$^1$}, Chaewon Kim$^1$\\
\noindent\textit{\textbf{Affiliations: }}\\
$^1$ Kookmin University, South Korea\\

\subsection*{MRT-ShadowR}
\noindent\textit{\textbf{Title: }}\\
Shadow Removal via Multi-stage Residual Transformer\\
\noindent\textit{\textbf{Members:}}\\ 
\textit{Alexandru Brateanu$^1$}, Ciprian Orhei$^2$, Cosmin Ancuti$^2$\\
\noindent\textit{\textbf{Affiliations: }}\\
$^1$ University of Manchester\\
$^2$ Politehnica University Timisoara\\

\subsection*{CV\_SVNIT}
\noindent\textit{\textbf{Title: }}\\
Deshadownet \\
\noindent\textit{\textbf{Members:}}\\ 
\textit{Tanmay Chaturvedi$^1$}, Manish Kumar$^1$, Anmol Srivastav$^1$, Daksh Trivedi$^1$, Shashwat Thakur$^1$, Kishor Upla$^1$\\
\noindent\textit{\textbf{Affiliations: }}\\
$^1$ Sardar Vallabhbhai National Institute of Technology\\

\subsection*{X-L}
\noindent\textit{\textbf{Title: }}\\
Efficient Hue Guidance Network for Single Image Shadow Removal\\
\noindent\textit{\textbf{Members:}}\\ 
\textit{Zeyu Xiao$^1$}, Zhuoyuan Li$^2$\\
\noindent\textit{\textbf{Affiliations: }}\\
$^1$ National University of Singapore\\
$^2$ University of Science and Technology of China\\

\subsection*{ZhouBoda}
\noindent\textit{\textbf{Members:}}\\ 
\textit{Boda Zhou}\\

\subsection*{GroupNo9}
\noindent\textit{\textbf{Members:}}\\ 
\textit{Shashank Shekhar$^1$}, Rishi Agarwal$^1$, Prachi Atri$^1$, Choudhary Sumit Jaswant$^1$, Prince Kumar$^1$, Shathak Arya$^1$, Sudhanshu Kumar$^1$, Kanka Dutta$^1$, Raushan Kumar$^1$\\
\noindent\textit{\textbf{Affiliations: }}\\
$^1$ Jawaharlal Nehru University \\

\end{document}